\documentclass[12pt]{article}

\usepackage{epsfig}
\usepackage{graphicx}
\usepackage{amsmath}
\usepackage{amssymb, verbatim}
\usepackage{multirow}
\usepackage{fullpage}

\begin{document}

\title{Estimating 3D Human Shapes from Measurements}
\author{Stefanie Wuhrer\footnote{National Research Council of Canada} \footnote{Saarland University, Germany} \footnote{Max-Plank Institute Informatik, Germany} \and Chang Shu$^*$}

\maketitle

\begin{abstract}
The recent advances in 3-D imaging technologies give rise to databases of human shapes, from which statistical shape models can be built. These statistical models represent prior knowledge of the human shape and enable us to solve shape reconstruction problems from partial information. Generating human shape from traditional anthropometric measurements is such a problem, since these 1-D measurements encode 3-D shape information. Combined with a statistical shape model, these easy-to-obtain measurements can be leveraged to create 3D human shapes. However, existing methods limit the creation of the shapes to the space spanned by the database and thus require a large amount of training data. In this paper, we introduce a technique that extrapolates the statistically inferred shape to fit the measurement data using non-linear optimization. This method ensures that the generated shape is both human-like and satisfies the measurement conditions. We demonstrate the effectiveness of the method and compare it to existing approaches through extensive experiments, using both synthetic data and real human measurements.
\end{abstract}

\section{Introduction}

Many applications require realistic 3D human shapes. For instance, 3D human models are used to design products that fit a target population. Typically, these shapes need to have certain characteristics or to be samples from a population. Although it is possible to digitize humans using 3D imaging technologies, it is impractical to find and scan suitable human subjects for each individual application. On the other hand, there is a long history of using anthropometric measurements to describe the human shape. These measurements are linear and curvilinear distances between anatomical landmarks or circumferences at predefined locations. In spite of providing limited shape information, they are readily available, and therefore widely used. 

Intuitively, the linear measurements should encode information about the body shape. If accurate 3D human shape can be inferred from a simple set of measurements, traditional anthropometric data can be leveraged to create 3D datasets without resorting to expensive scanning equipment.

The early attempts at solving this problem deform a generic human model to fit the measurement data. However, the models thus obtained are not necessarily human forms because generating a human shape from a sparse set of measurements is an under-constrained problem. More sophisticated methods use knowledge of the human shape learned from a database of 3D scans. A statistical model can be built from a set of registered 3D scans and a relationship between the measurements and the shape space can be established. In doing so, detailed 3D shapes can be predicted from sparse and partial shape data. Apart from predicting 3D shapes from measurements~\cite{seo_magnenat-thalmann_03, allen_curless_popovic_03_parametrization_body_shape}, other partial shape information, such as marker positions~\cite{anguelov_srinivasan_koller_thrun_rodgers_05_shapecomp} and 2D images~\cite{guan_etal, chen_cipolla_SPGLVM_reconstruction}, has also been used to reconstruct 3D shapes.

Existing methods that predict the 3D shapes using a statistical model limit the generated shapes to 
the space spanned by the shape model (more specifically, within two standard deviations of the shape variability) and local variations of human shapes that are outside this space cannot be predicted accurately. This means that these methods require a 3D database that accurately represents a target population (for instance, human subjects of different ethnicities and different weight classes must be present in the database). To apply these techniques, one must carefully choose a target population and acquire a representative database. In reality, the acquisition of a 3D anthropometry data that represents a target population well is complicated and expensive (the acquisition of the Civilian American and European Surface Anthropometry Resource took 4 years and cost \$6 Million\footnote{http://www.sae.org/standardsdev/tsb/cooperative/caesumm.htm}). 

In this paper, we address the problem of estimating 3D human body and face shapes given a set of anthropometric measurements. Our main goal, in contrast to previous work, is to predict shapes that are inside the space of human shapes, but still account for local variations not captured by the training data. This extrapolation allows us to create human shapes based on relatively small 3D databases.

As most of the previous approaches, our method does not take posture or expression changes into account. That is, we assume that all of the training data is in a standard pose. This scenario is commonly assumed since in a typical 3D anthropometry survey, the human subjects are asked to maintain a standard posture.

The rest of the paper is organized as follows. Section~\ref{Approach} gives an overview of the approach, and Section~\ref{Refinement} outlines the details of a shape refinement step, which constitutes the main contribution of this paper. We aim for design applications in engineering where accuracy is essential. We show, through experiments, that straightforward application of existing techniques does not produce satisfactory results (Section~\ref{Experiments}).

\section{Related Work}
\label{Related}

This section reviews existing work that aims to estimate 3D human shapes based on partial input information. We categorize the related literature by the type of input data that is used to predict human shapes.

\subsection*{Estimating 3D Body or Face Shapes from Measurements}
DeCarlo et al.~\cite{decarlo_etal_98} presented an early attempt at using anthropometric measurements to compute human face shapes. The method generates a random set of measurements based on a set of rules that capture, for instance, typical facial proportions. To create a 3D face shape, the method deforms a template model to fit these measurements using a free-form deformation technique called variational modeling. Variational modeling allows to give the measurements as constraints when deforming the template. To ensure that the estimated shape remains in the shape space of human shapes, much care is taken when creating the measurements that are used as constraints. While this method generates human-like face shapes using a statistical model, the accuracy of the shapes created by this method is limited by the sparse set of measurements used to represent the training data. That is, the results are overly smooth and do not contain realistic details. It is not straight forward to extend this technique to allow for the generation of human body shapes from an arbitrary set of anthropometric measurements. By combining statistical learning with a mesh-based technique, our method allows the generation of realistic and detailed human body shapes from an arbitrary set of anthropometric measurements.

The following approaches estimate human body shapes from a discrete set of measurements on the body, all proceeding by learning a linear or near-linear mapping between the training set of human shapes and the parameter space of measurements and then using this mapping to predict a shape based on a new set of measurements. While these approaches work well when the predicted shape is inside the shape space spanned by the training set, they do not allow for extrapolations from this shape space. 

Wang~\cite{Wang2005} uses a parameterized database of human shapes consisting of \textit{feature patches} to find a new human shape based on a given set of measurements. Feature patches are initialized to smooth patches that are subsequently refined to capture geometric details of the body shape. While this refinement does not maintain the smoothness of the patches, the final results are visually smooth and do not contain realistic localized shape details. The approach finds models in the database with measurements similar to the given set of measurements and computes the new human shape as a linear combination of these models. Wei et al.~\cite{wei_etal_08} use a similar approach, but the models are represented using a layer-based representation. 

Allen et al.~\cite{allen_curless_popovic_03_parametrization_body_shape,allen_curless_popovic_04} compute a new triangular mesh based on a given set of measurements. Starting from a parameterized database of human bodies in similar poses, the approach performs Principal Component Analysis (PCA) of the data. This yields one PCA weight for each training shape. The training database is used to learn a linear mapping from the set of measurements measured on the training data to the PCA space. This mapping is called \textit{feature analysis}. Feature analysis can be used to compute a new PCA weight based on a new set of measurements, and the learned PCA model allows to compute a new triangular mesh from this PCA weight. Chu et al.~\cite{chu_etal_2010} perform feature analysis on a database of human shapes consisting of feature patches to find a new model consisting of smooth patches. 

Seo and Megnenat-Thalmann~\cite{seo_magnenat-thalmann_03} represent the human body as a triangular mesh with an associated skeleton model. As with the approach of Allen et al., this approach reduces the dimensionality of the data using PCA. This yields a set of PCA weights. The approach learns a mapping from the set of measurements measured on the training data to the PCA space using an interpolating radial basis function (RBF) with Gaussian kernel~\cite{dryden_mardia_shape_analysis}. As in the approach of Allen et al., this mapping produces a shape based on a new set of measurements. Hasler et al.~\cite{HasStoSunRosSei09} apply the two previously reviewed approaches to a new representation of human body shapes that is posture invariant. Their method simultaneously models body pose and shape. 

Recently, Baek and Lee~\cite{Baek2010} presented a technique that uses hierarchical clustering to build a statistical model of the training data. The approach proceeds by clustering the training database and by performing a multi-cluster analysis of the training data. To predict a body shape based on a set of input measurements, the approach finds the shape within the learned shape space that best describes the measurements using an optimization of shape parameters. It is shown experimentally that accurate and visually pleasing body shapes are estimated when the input body sizes are inside the shape space spanned by the training data. This approach is conceptually similar to the first optimization step of our algorithm. Hence, we expect that this approach does not allow to model shape variations that are outside of the shape space spanned by the training data.

\subsection*{Estimating 3D Body or Face Shapes from Markers}
Anguelov et al.~\cite{anguelov_srinivasan_koller_thrun_rodgers_05_shapecomp} aim to estimate a 3D human body shape based on a sparse set of marker positions. 
This technique is useful when motion capture data is available. Anguelov et al.'s SCAPE model represents the human body as a triangular mesh with an associated skeleton model. As with the approach of Allen et al., this approach reduces the dimensionality of the data using PCA. This yields a set of PCA weights. Anguelov et al. aim to compute a new triangular mesh based on a set of marker positions located on the body. This is achieved by adjusting the PCA weights to solve a non-linear optimization problem. This method searches the solution in the learned PCA space. Hence, it cannot find local variations not present in the training database.

\subsection*{Estimating 3D Body or Face Shapes from Images}
Finally, we review approaches that aim to estimate human body shapes based on a set of input images. The following approaches proceed by learning a correlation between a training set of 3D face or body shapes and a set of derived 2D images and by using this learned correlation to predict a shape based on a new set of 2D images. These approaches work well when the predicted shape is inside the shape space spanned by the training set. However, they do not handle optimizations outside the learned shape space of 3D models.

Blanz and Vetter~\cite{blanz_vetter_99} estimate a 3D face shape from a single input image in neutral expression. They start by building a parameterized database of textured 3D faces and performing PCA on the shape and texture data. Given an input image, the learned PCA space is searched to find the textured shape (and parameters related to rendering the model) that best explains the input image.

Seo et al.~\cite{seo_etal_shape_from_silhouette} estimate a body shape from two images of a human in a fixed posture. Starting from a parameterized database of human meshes in similar poses, the approach performs PCA of the 3D data. Given the two images, the learned PCA space is searched to find a set of PCA weights that corresponds to a 3D shape that matches the input images well. 

Chen and Cipolla~\cite{chen_cipolla_SPGLVM_reconstruction} aim to estimate the human body shape in a fixed pose based on a given silhouette. Starting from a parameterized database of human meshes in similar poses and a set of corresponding silhouettes, the approach performs PCA of the 3D and 2D data separately. The approach then computes a mapping from the PCA space of the silhouette data to the PCA space of the 3D data using a Shared Gaussian Process Latent Variable Model (SGPLVM)~\cite{shon_grochow_hertzmann_rao_SGPLVM}. Given a new silhouette, the approach maps the silhouette into silhouette PCA space and uses the SGPLVM to map to the PCA space of the 3D meshes. Ek et al.~\cite{ek_torr_lawrence_07} use a similar approach to estimate the pose of a human body based on a given silhouette.

Guan et al.~\cite{guan_etal} estimate both the shape and pose of a human body shape from a single photograph with a set of markers to be identified by the user. The approach is based on the SCAPE model. When adjusting the PCA weights, the shape is deformed to best match the image in terms of a shape-from-shading energy. Hasler et al.~\cite{hasler_ackermann_etal_10} estimate both the shape and pose of a human body from a photograph of a dressed person. This approach requires a manual segmentation of the background and the human in the image.

\section{Approach}
\label{Approach}

This section outlines the proposed approach. As input to the method, we are given a database of triangular manifold meshes $X_0, \ldots, X_{n-1}$ of human bodies or faces with similar posture or expression and a set of measurements $\mathcal{S}$ to be considered. Let $P_i$ in $\mathcal{S}$ denote the measurements corresponding to $X_i$. Furthermore, we are given a set of distances $P_{new}$. Our aim is to estimate a shape $X_{new}$ that interpolates the distances $P_{new}$.

As previous methods, the approach proceeds by learning a correlation between the shapes and the measurements. When predicting a new shape, our approach finds an initial solution based on the learned correlation. Unlike previous approaches, however, our approach refines this solution to fit the measurements using two steps of non-linear optimization. First, we optimize the shape of the model with respect to the learned shape space. That is, we aim to find the point in the learned shape space that best describes the measurements. This gives a realistic human shape drawn from the distribution fitted to the training data. Hence, this shape can only contain local shape variations present in the training data. To account for other shape variations, we perform a second mesh-based optimization. This optimization deforms the shape to fit the measurements as close as possible while satisfying a smoothness constraint without using prior knowledge of the body shape, and therefore predicts shapes with local variations not present in the training database. 

When using a mesh-based deformation that optimizes the sought measurements, we can only expect to obtain a realistic human body shape if we start with a body shape that is close to the solution. For this reason, we start from the shape in the learned shape space that best describes the sought measurements as opposed to starting directly from the point in shape space predicted by the learned correlation. 

The following sections describe the details of the approach.

\subsection{Measurements}
\label{sec_measurements}

Anthropometric measurement represents a valuable source of human shape information, from which we can infer the 3D shapes for design applications. To learn a relationship between the anthropometric measurements and a set of 3D models, the measurements computed from the models should be equivalent to the measurements conducted by a human operator. The measurements can be one of the three types: Euclidean, geodesic, and circumference. An Euclidean or geodesic distance can be easily computed from two vertices on the model. 
A circumference can be computed by intersecting (part of) the model with a plane, finding the (possibly closed) polygonal chain of the intersection that contains a specified vertex, and measuring the circumference of the convex hull of this chain. This type of measurement is shown for the hip and waist circumferences in Figure~\ref{measurements_many_human}. The reason we measure the length of the convex hull instead of the length of the chain itself is that many anthropometric measurements, such as the chest circumference, measure the length of a convex hull. We specify the intersecting plane $\pi$ using the specified vertex $p$ and a normal direction $\vec{n}$. Furthermore, we specify a part of the model that is to be intersected with the plane $\pi$. This is necessary for some measurements, such as for the chest or hip circumference, where we wish to only intersect the torso (and not the arms) of the model with $\pi$. Note that the intersection of $\pi$ with a subset of $X$ consists of a set of polygonal chains, where every vertex of the chain is either a vertex of $X$ or the intersection of an edge of $X$ with $\pi$. 

\subsection{Training}

We now consider learning a correlation between the set of shapes $X_i$ and the corresponding measurements $P_i$. 
To do this, the database of the human shapes first needs to be parameterized, a process that computes point-to-point correspondences among the shapes. This is in general a difficult problem~\cite{vanKaick_egstar10}. In practice, anthropometric markers identifying salient anatomic positions are often used to guide correspondence process. Such markers are provided in 3D anthropometric surveys, for example, the Civilian American and European Surface Anthropometry Resource (CAESAR) database~\cite{robinette_daanen_paquet_99_caesar}. In this work, the known marker positions are used to deform a template shape to each subject of the database~\cite{allen_curless_popovic_03_parametrization_body_shape, xi_lee_shu_07_bodies}. 

The Euclidean and geodesic measurements are specified by their two endpoints and the type of distance to be considered and the circumference measurements are given by one plane (specified by a vertex and a normal vector) and the part of the body to be intersected with that plane. Since the database is parameterized, the vertices and body parts can be given in terms of their vertex and triangle numbers on the mesh. 

At this point, a set of distances $P_0, \ldots, P_{n-1}$ can be computed for $X_0, \ldots, X_{n-1}$. We then learn a mapping between $\mathcal{S}$ and the space of human body shapes using feature analysis. To this end, we first perform PCA on the meshes $X_i$. In PCA space, each shape $X_i$ is represented by a vector $W_i$ of PCA weights. PCA yields a mean shape $\mu$ and a matrix $A$ that can be used to compute a new shape $X_{new}$ based on a new vector of PCA weights $W_{new}$ as $X_{new} = A W_{new} + \mu$. Recall that the aim is to create a new shape based on a new point $P_{new}$ in $\mathcal{S}$. To achieve this goal, feature analysis learns a linear mapping from $P_i$ to $W_i, i=0,\ldots,n-1$. This yields a matrix $B$ that can be used to compute a new vector of PCA weights $W_{new}$ based on a new point $P_{new}$ as $W_{new}=B P_{new}$, which generates a shape $X_{new}$ based on a new point $P_{new}$ as $X_{new} = A B P_{new} + \mu$. To give all PCA coordinates the same weight, we normalize each entry of the PCA weights by its corresponding PCA eigenvalue before performing feature analysis. 

\subsection{Prediction}

The mapping between the measurements and the PCA weights of the 3D shapes learned in the previous section allows us to find an initial shape $X_{new}^{init}$ given a new set of measurements $P_{new}$.
However, straightforward application of feature analysis has the disadvantage that the shapes obtained for atypical $P_{new}$ are not contained in the space of human shapes (e.g. see Figure~\ref{atypical_prediction}). Since the subsequent refinement process depends on this initial shape, it is important to restrict it to be within the range of human shape.

Given the PCA weights obtained by feature analysis $W_{new}=B P_{new}$, we
restrict $W_{new}$ such that each dimension $W_{new}[i]$ is at most $l$ times the standard deviation of the PCA space along dimension $i$. This choice is based on the assumption that the shape space is modeled as independent Gaussian distributions and most of the shapes are located within $l$ standard deviations of the mean for a suitable parameter $l$. Hence, this step restricts the reconstructed shape to stay within the space of human shapes that was learned using PCA. In our implementation, we use the normalized weight vector to find an initial shape $X_{new}^{init}$. 

The shape generated from the PCA weights can only serve as an initial estimation, because the shape space spanned by the PCA is limited by the sample size and may not account for all of the human shape variation. To find a shape that respects the required measurements while staying in the learned shape space, we need to refine $X_{new}^{init}$. The following section presents our novel method for shape refinement in detail.

\section{Shape Refinement}
\label{Refinement}

This section outlines a novel approach to refine the initial estimate $X_{new}^{init}$ to respect the required measurements while staying in the learned shape space.
We formulate the refinement problem as an energy minimization problem. 

Let $p_0, \ldots, p_{m-1}$ denote the vertices of $X_{new}$ and let $\vec{p}_0, \ldots, \vec{p}_{m-1}$ denote their position vectors. Furthermore, let $W_{new}=A^+(X_{new}-\mu)$ be the vector of PCA weights corresponding to $X_{new}$, where $A^+$ is the pseudo-inverse of $A$. 

For Euclidean measurements, we are given the target length $l_t(d)$ of the Euclidean distance of the segment $d$ between two vertices $p_i$ and $p_j$. The shape $X_{new}$ that satisfies all given Euclidean distance constraints minimizes
$$E_e = \sum_{d \in \mathcal{D}}((\vec{p_i}-\vec{p_j})^2 - (l_t(d))^2)^2,$$
where $\mathcal{D}$ is the set of all Euclidean distances that have a desired length, and $p_i$ and $p_j$ are the endpoints of $d$.

For geodesic measurements, we are given the target length $l_t(P)$ of the geodesic path $P$ between two vertices $p_i$ and $p_j$. Let $l_g(P)$ denote the geodesic length of $P$. In the following, we assume that the relative length of each edge $e$ of $P$ with respect to $l_g(P)$ does not change during the deformation of the mesh. This is a reasonable assumption as we wish to preserve relative edge lengths during the deformation. Furthermore, this assumption translates into an easy optimization problem. With this assumption, we can compute the target length $l_t(e)$ of $e$ as $l_t(e) = \frac{l_t(P)}{l_g(P)}l_g(e)$, where $l_g(e)$ is the current length of $e$. Then, the shape $X_{new}$ that satisfies all given geodesic distance constraints minimizes
$$E_g = \sum_{e \in \mathcal{P}}((\vec{p_k}-\vec{p_l})^2 - (l_t(e))^2)^2,$$
where $\mathcal{P}$ is the set of all geodesic paths that have a desired length, and $p_k$ and $p_l$ are the endpoints of $e$.

For circumference measurements, we are given the target length $l_t(C)$ of the circumference $C$ of the convex hull of the polygonal chain passing though vertex $p_i$ and contained in the intersection between the mesh and the plane $\pi$ though $p_i$ with normal direction $\vec{n}$. We compute the length $l_g(C)$ of the circumference on the mesh as outlined in Section~\ref{sec_measurements}. Let $q_i$ denote all of the points on the convex hull of the polygonal chain. With a similar reasoning as above, in the following, we assume that the relative length of each edge $e$ of the convex hull with respect to $l_g(C)$ does not change during the deformation of the mesh. Hence, we can compute the target length $l_t(e)$ of $e$ as $l_t(e) = \frac{l_t(C)}{l_g(C)}l_g(e)$, where $l_g(e)$ is the current length of $e$. Then, the shape $X_{new}$ that satisfies all given circumference constraints minimizes
$$E_c = \sum_{e \in \mathcal{C}}((\vec{q_i}-\vec{q_j})^2 - (l_t(e))^2)^2,$$
where $\mathcal{C}$ is the set of all circumferences that have a desired length, and $q_i$ and $q_j$ are the endpoints of $e$. Recall that $q_i$ and $q_j$ are not necessarily vertices of the mesh.

Our aim is to deform $X_{new}^{init}$, such that $E_m = E_e+E_g+E_c$ is minimized. We solve this problem using two consecutive steps by minimizing $E_m$ with respect to $W_{new}$ followed by minimizing $E_m$ with respect to $\vec{p}_i$. The first minimization finds the shape in the learned shape space that best describes the measurements. This step finds a realistic human body shape whose measurements are close to the sought ones. The second minimization then further deforms the shape using a mesh-based optimization without using prior knowledge on the shape. This minimization allows the shape to deform locally in ways not present in the training data. During this minimization, a smoothness term is used to keep a realistic body shape.

\subsection{Minimization with respect to $W_{new}$} 
First, we discuss how to minimize $E_m$ with respect to $W_{new}$. This step finds the shape in the learned shape space that best describes the measurements. We solve the optimization problem using a quasi-Newton approach. This approach has the advantage of having a near-quadratic convergence rate if the initial solution is close to the minimum. Since this approach requires analytic gradients, we next discuss the derivatives of the different energy terms.

The derivative of $E_e$ with respect to $\vec{p_i}$ is 
$$\nabla_{\vec{p_i}} E_e = \sum_{d \in D(p_i)} 4((\vec{p_i}-\vec{p_j})^2 - (l_t(d))^2)(\vec{p_i}-\vec{p_j}),$$
where $D(p_i)$ is the set of distances in $D$ with endpoint $p_i$. The derivative of $E_e$ with respect to $W_{new}$ is $\nabla_{W_{new}} E_e = A^+ \nabla_{\vec{p_i}} E_e$. The derivatives of $E_g$ with respect to $\vec{p_i}$ and $W_{new}$ are similar to $\nabla_{\vec{p_i}} E_e$ and $\nabla_{W_{new}} E_e$. To compute the derivative of $E_c$ with respect to $\vec{p_i}$, recall that each vertex $q_i$ of the convex hull of the polygonal chain is either a vertex of the mesh or the intersection of an edge of the mesh with $\pi$. Hence, every $q_i$ can be expressed as a convex combination of at most two vertices of the mesh. Using this formulation allows us to compute the derivatives of $E_c$ with respect to $\vec{p_i}$ and $W_{new}$ in a similar way to $\nabla_{\vec{p_i}} E_e$ and $\nabla_{W_{new}} E_e$.

Note that the geodesic path between $p_i$ and $p_j$ and the circumference measurement specified by $p_i$ and $\vec{n}$ may change when the mesh is deformed. To remedy this problem, we update the geodesic paths and the circumference measurements and solve the resulting optimization problem $s$ times. This finds the point $W_{new}$ in PCA space corresponding to the shape that has measurements closest to the desired ones. As before, we normalize $W_{new}$ such that each dimension $W_{new}[i]$ of $W_{new}$ is at most $l$ times the standard deviation of the PCA space along dimension $i$ to ensure that the shape stays within the learned shape space. Finally, $W_{new}$ is used to compute the shape $X_{new}^{pca}$ as $X_{new}^{pca} = A W_{new}+\mu$.

\subsection{Minimization with respect to $\vec{p}_i$} 
Second, we minimize $E_m$ with respect to the vertex positions $\vec{p}_i$. This term ensures that local variations not present in the training data are predicted correctly by locally deforming the shape to fit the required measurements. However, simply minimizing $E_m$ may result in meshes that are not smooth. Hence, we encourage close-by parts of the mesh to deform similarly by considering the smoothness energy 
$$E_s = \sum_{p_i \in X_{new}} \sum_{p_j \in N(p_i)} (\Delta \vec{p_i}- \Delta \vec{p_j})^2,$$
where $\Delta \vec{p_i}$ is the translation vector by which $\vec{p_i}$ is moved and $N(p_i)$ is the one-ring neighborhood of $p_i$. The derivative of $E_s$ with respect to $\vec{p_i}$ is 
$$\nabla_{\vec{p_i}} E_s = \sum_{p_j \in N(p_i)}2(\Delta \vec{p_i}- \Delta \vec{p_j}).$$
Note that the smoothness term helps ensure the predicted shape stays withing the shape space of human shapes. The vertex positions are initialized to the vertex positions of $X_{new}^{pca}$. We minimize the combined energy $E = (1-\lambda) E_m + \lambda E_s$ with respect to $\vec{p_i}$, where $\lambda$ is the weight for the smoothing term. As before, we update the geodesic paths and the circumference measurements and solve the resulting optimization problem $s$ times, yielding the final result $X_{new}$.

\subsection{Discussion of parameters} 
The proposed method depends on the parameters $l, \lambda, s$, and on the termination criteria of the energy minimization. In our experiments, we stop the energy minimization if the relative change of the energy is smaller than $1e8$ times the machine accuracy, if the norm of the gradient is smaller than $1e-8$, or if 100 iterations have been performed. The parameter $l$ determines the size of the acceptable shape space for reconstruction. This parameter provides a way to trade off high measurement accuracy and likely human shapes. If the measurements given to the method are known to be accurate measurements of a real human, $l$ can be set to a high value. If the measurements given to the method are randomly generated, however, $l$ needs to be low to avoid non-human shapes. Hence, in our experiments with real data, we set $l=10$ and in our experiments with synthetic data, we set $l=3$. The parameter $\lambda$ determines the relative weight of the smoothing term. In our experiments, we set $\lambda=0.1$. The parameter $s$ determines how often the geodesic paths and circumferences are recomputed. We need to set $s$ large enough to achieve the required accuracy, yet small enough to obtain an efficient approach. In our experiments, we set $s=3$ unless specified otherwise. 

We leave it for future work to find the optimal parameter settings automatically. This is a challenging problem because the result depends on the parameter values, and because including the parameter values in the set of parameters to be optimized gives a highly nonlinear optimization problem. 
 
\section{Experiments}
\label{Experiments}

This section demonstrates the effectiveness of our method using both synthetic experiments and experiments with real data. 

We implemented the proposed algorithm using C++. The implementation uses Dijkstra's algorithm~\cite{Dijkstra1959} to compute geodesic distances, the quickhull algorithm~\cite{Barber96thequickhull} to compute the convex hull to determine circumferences, and the limited-memory Broyden-Fletcher-Goldfarb-Shanno scheme~\cite{liu_nocedal_lbfgsb} to solve the optimization problems. We choose these standard algorithms for their ease of use. To parameterize the database used for training, the efficient approach by Xi et al.~\cite{xi_lee_shu_07_bodies} is employed. 

We compare our results to the results obtained using feature analysis and to the results obtained by using a SGPLVM mapping from the space of the measurements to the PCA space of the models. Both approaches are reviewed in Section~\ref{Related}. We use the code by Ek et al.~\cite{ek_torr_lawrence_07} to compute the SGPLVM.

\subsection{Input Measurements}

The synthetic experiments aim to reconstruct human face and body shapes. For each synthetic experiment, we learn a multivariate Gaussian distribution $\mathcal{N}(\mu, \Sigma)$ from the available sets of measurements. First, 200 sets of measurements are generated by randomly sampling 200 points from $\mathcal{N}(\mu, \Sigma)$. Note that since these samples are drawn from the learned distribution, most of the measurements are similar to the ones in the training set. We denote this set of measurements by $S_{close}$.
Second, 200 sets of measurements are generated by randomly sampling 200 points that are located on the ellipsoidal surface $x^T \Sigma^{-1} x$, where $x = \mu + k \Sigma_d$, $k$ is a constant, and $\Sigma_d$ is the vector that contains the diagonal elements of $\Sigma$. Note that the larger the $k$ is, the farther the measurements are from the training set. Let $S_{k}$ denote this set of measurements. In our experiments, we set $k=2,4$.

The experiments based on real data aim to reconstruct human body shapes from a set of measurements. We consider two types of input data. For the first type of input data, we digitally measure the distances in $\mathcal{S}$ on a number of parameterized 3D human body scans and use these measurements as input to our algorithm. We denote these measurements by $S_{real}$. Note that in this case, the ground truth of the 3D body shape is known and can be used to evaluate the results.

For the second type of input data, we asked a number of volunteers to measure each other. This results in a set of real-world measurements of real subjects. We denote these measurements by $S_{meas}$. As in this case, no ground truth 3D body shape is known, we validate the accuracy of the 3D reconstruction visually by comparing the 3D shape to silhouettes of photographs of the volunteers. This experiment shows the accuracy we can expect when a person who is not an expert in anthropometry tries to use our method to build a digital clone of herself.

\subsection{Face Experiments}

This section validates our approach using a set of synthetic experiments on 3D human face shapes.

\subsubsection*{Training Data}

\begin{figure}[htb]
\centering
\includegraphics[width = 2.5cm]{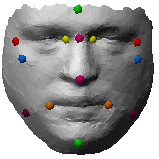}
\caption{\textit{Measurements used to define shape space for the face experiments.}}
\label{measurements}
\end{figure}

The training set consists of 50 faces from the CAESAR data\-base~\cite{robinette_daanen_paquet_99_caesar}. After parameterization, each face contains 6957 triangles. This experiment considers the seven measurements shown in Figure~\ref{measurements}. Here, each dimension measures the geodesic distance between the pair of points shown in the same color. Note that several geodesic paths may overlap. 

\subsubsection*{Synthetic Experiments}

We generate new face shapes that aim to interpolate the measurements in $S_{close}$ and $S_{k}$ using our approach, feature analysis, and SGPLVM. Table~\ref{error_table} shows the errors of the generated faces for each method. We define an error for each dimension as the absolute value between the distance on the generated shape and the input measurement for this dimension, and use this to compute the error in mm as the average error over all dimensions. Note that feature analysis yields the lowest average errors and that SGPLVM yields the highest average errors for all experiments. 

\begin{table}[htb]
\centering
\small
\begin{tabular}{|c||r|r|r|}
\cline{1-4} 
& $S_{close}$ &  $S_{2}$ & $S_{4}$\\
\cline{1-4} 
Feature Analysis & 0.67& 1.63& 6.76\\
\cline{1-4} 
SGPLVM & 5.35& 5.28& 10.88\\
\cline{1-4} 
Our Approach &  0.68& 2.00& 7.27\\
\cline{1-4} 
\end{tabular}
\caption{\it{Comparison between our approach, feature analysis, and SGPLVM for the synthetic face experiments. The error in mm is the average error over all dimensions.}}
\label{error_table}
\end{table}

\begin{figure}[htb]
\centering
\includegraphics[width = 6.0cm]{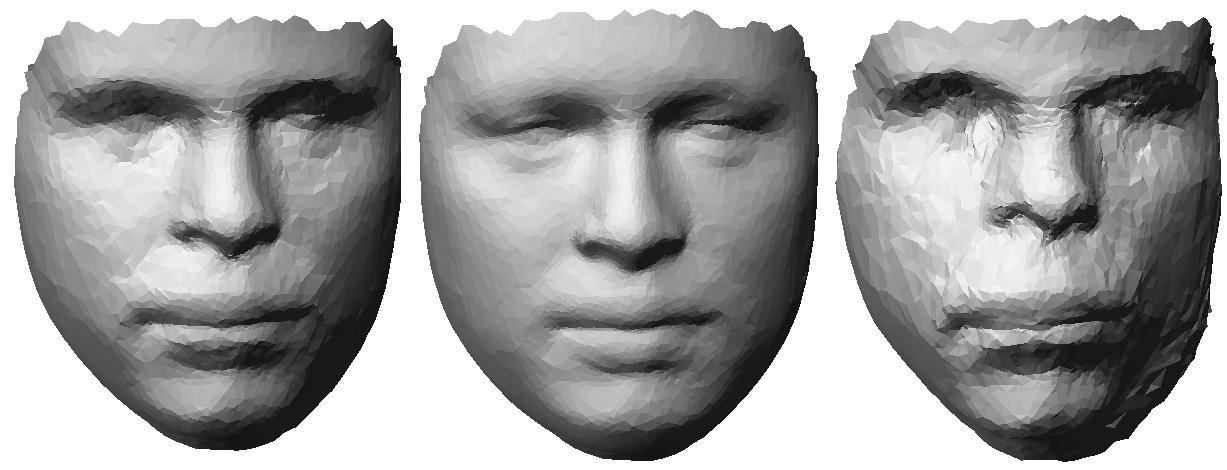}
\caption{\it{Face shape predictions from a random sample in $S_{4}$. Estimates from left to right using our approach, SGPLVM, and feature analysis.}}
\label{reconstructed_shapes}
\end{figure}

Several of the shapes predicted using feature analysis are clearly outside of the shape space of human faces, as shown on the right of Figure~\ref{reconstructed_shapes}. This is undesirable because human perception is sensitive with respect to unrealistic variations in face shape. In fact, most of the faces predicted using feature analysis in $S_{4}$ are clearly outside of the shape space of human faces. 

It is desirable to have variability among the computed shapes. Both feature analysis and our approach predict significantly different shapes for all of the experiments. SGPLVM produces slightly different face shapes for different samples of $S_{close}$ and $S_{2}$, but for all samples in $S_{4}$, the same face shape (visually similar to the mean shape) is produced. 

In summary, our approach is the only one of the compared approaches that yields a variety of face shapes while always estimating shapes that are in the space of human faces. Furthermore, our approach yields lower errors than SGPLVM.

\subsection{Body Experiments}

This section validates our approach using a set of synthetic experiments and experiments based on real data.

\subsubsection*{Training Data}

\begin{figure}[htb]
\centering
\includegraphics[height = 8.0cm]{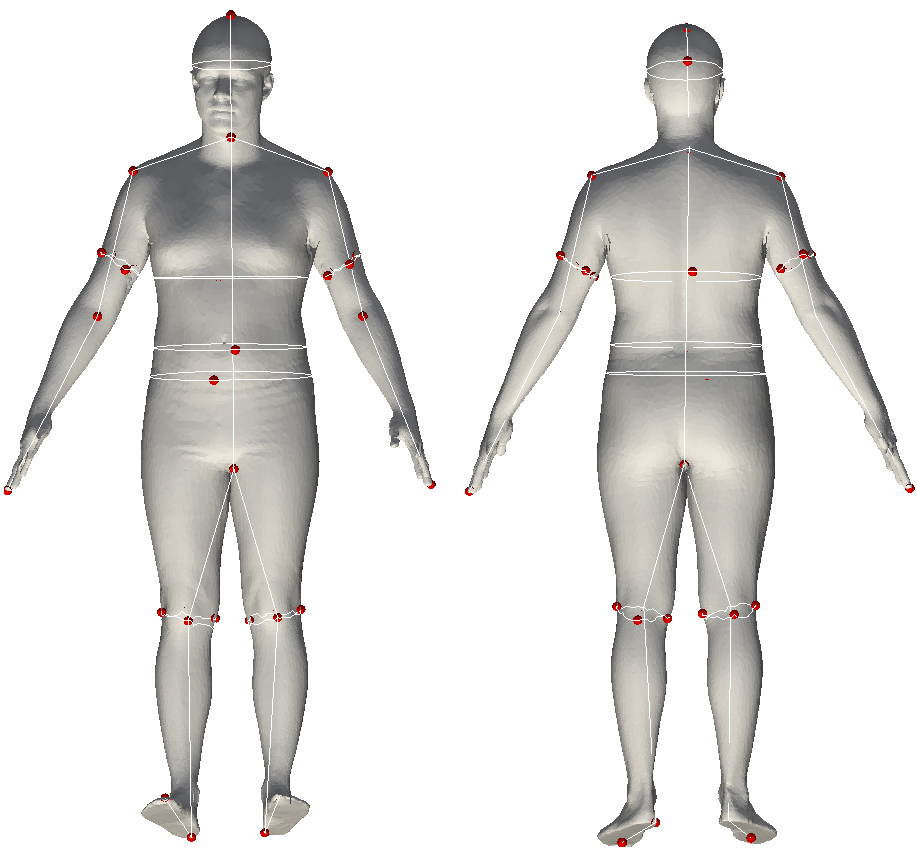}
\caption{\textit{Measurements used to define shape space for the body experiments. Red spheres show the specified points of the measurements and red lines show the measurements for this specific shape.}}
\label{measurements_many_human}
\end{figure}

The training set consists of 360 bodies of the CAESAR database. After parameterization, each body contains 60000 triangles. This experiment considers 34 measurements: 14 Euclidean measurements, 4 circumference measurements (hip, waist, chest, and head circumferences) and 4 circumferences defined by four geodesic measurements each (knee and arm circumferences). Figure~\ref{measurements_many_human} shows the measurements on one shape. 

\subsubsection*{Synthetic Experiments}

We generate new body shapes that aim to interpolate the new measurements in $S_{close}$ and $S_{k}$ using our approach, feature analysis, and SGPLVM. Table~\ref{error_table_body} shows the errors of the generated bodies for each method. The errors are computed in the same way as in the previous experiment. Note that our approach yields the lowest average errors for all experiments. 

\begin{table}[htb]
\centering
\small
\begin{tabular}{|c||r|r|r|r|}
\cline{1-5} 
& $S_{close}$ &  $S_{2}$ & $S_{4}$ & $S_{real}$ \\
\cline{1-5} 
Feature Analysis & 3.29& 14.26& 36.86& 3.09 \\
\cline{1-5} 
SGPLVM & 4.91& 12.19& 22.88& 4.55 \\
\cline{1-5} 
Our Approach &  1.74& 6.39& 15.49& 1.10 \\
\cline{1-5} 
\end{tabular}
\caption{\it{Comparison between our approach, feature analysis, and SGPLVM for the body experiments. The error in mm is the average error over all dimensions.}}
\label{error_table_body}
\end{table}

\begin{figure*}[htb]
\centering
\includegraphics[width = \textwidth]{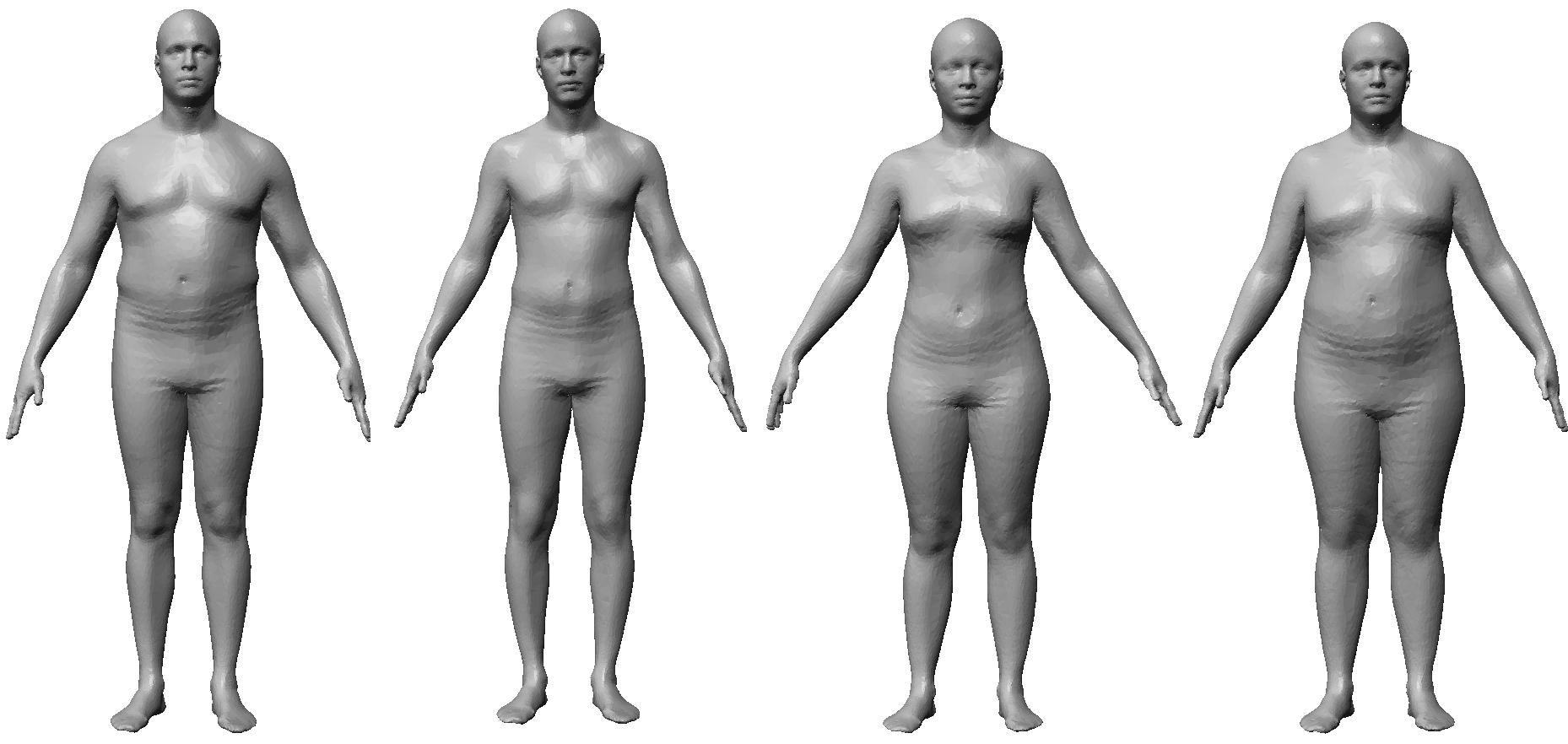}
\caption{\it{Four of the shapes computed using our approach from random samples in $S_{close}$.}}
\label{reconstructed_shapes_body}
\end{figure*}

For most samples in $S_4$ and for several samples in $S_2$, feature analysis predicts shapes that are outside of the shape space of human bodies. For the samples in $S_{close}$, feature analysis yields visually pleasing shapes. SGPLVM always yields body shapes that are in the shape space of human bodies. However, for the samples in $S_4$, the variation of the predicted body shapes is low. Some of the shapes predicted using $S_4$ with our approach contain small local artifacts due to the optimization with respect to the vertex coordinates. Note that setting $\lambda$ to a higher value will reduce these artifacts at the cost of increased error. For $S_2$ and $S_{close}$, our approach always predicts globally and locally realistic body shapes. Figure~\ref{reconstructed_shapes_body} shows four of the body shapes in $S_{close}$ predicted using our approach. We can see that a large variety of shapes can be computed.

\subsubsection*{Experiments Using Digital Measurements of Real Subjects}

For this experiment, we consider 50 additional parameterized body shapes from the CAESAR database and use each of these models to digitally measure the 34 relevant measurements. We call this set of measurements $S_{real}$. 

\begin{figure*}[htb]
\centering
\begin{tabular}{c c c c} 
\includegraphics[height = 6.0cm]{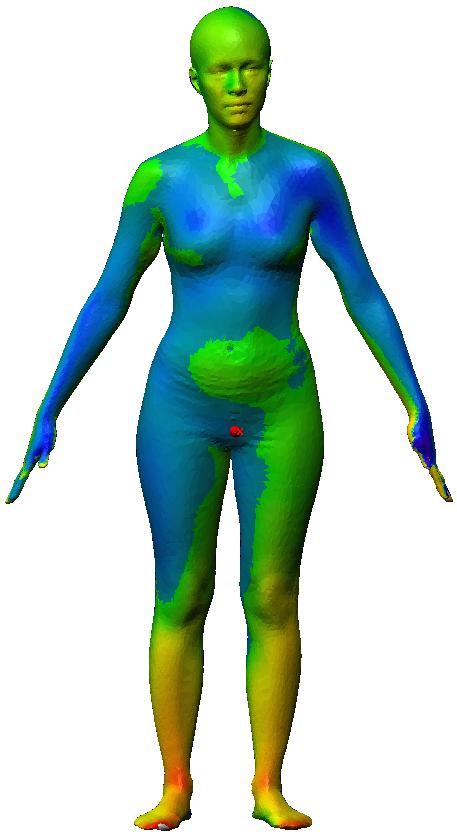} & 
\includegraphics[height = 6.0cm]{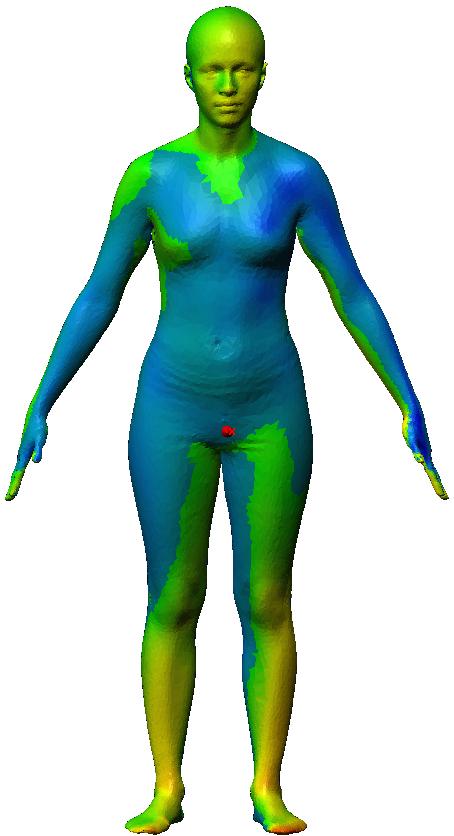} & 
\includegraphics[height = 6.0cm]{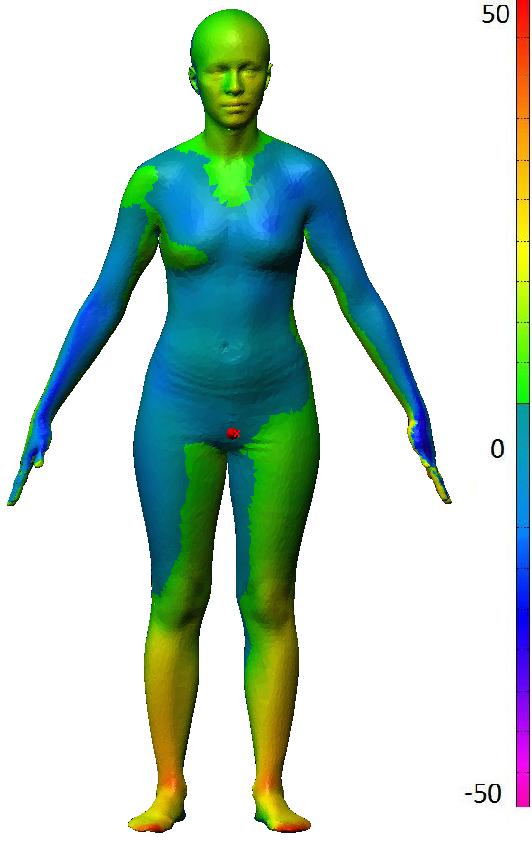} &
\includegraphics[height = 6.0cm]{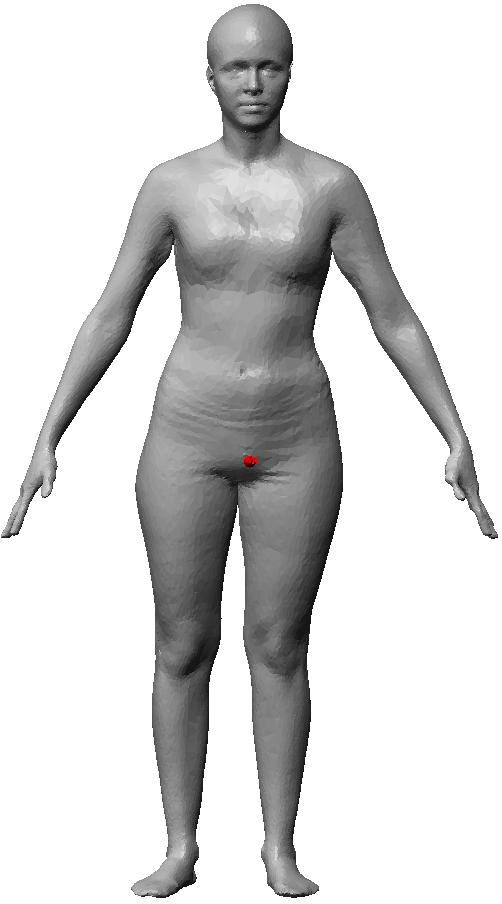} \\
\includegraphics[height = 6.0cm]{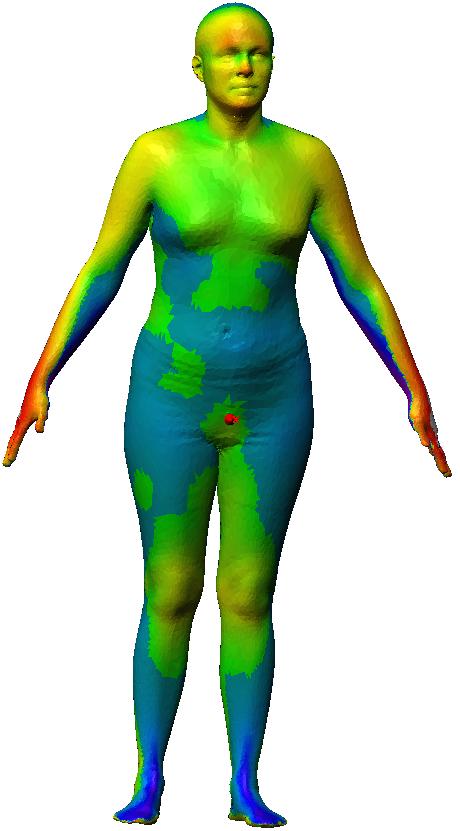} & 
\includegraphics[height = 6.0cm]{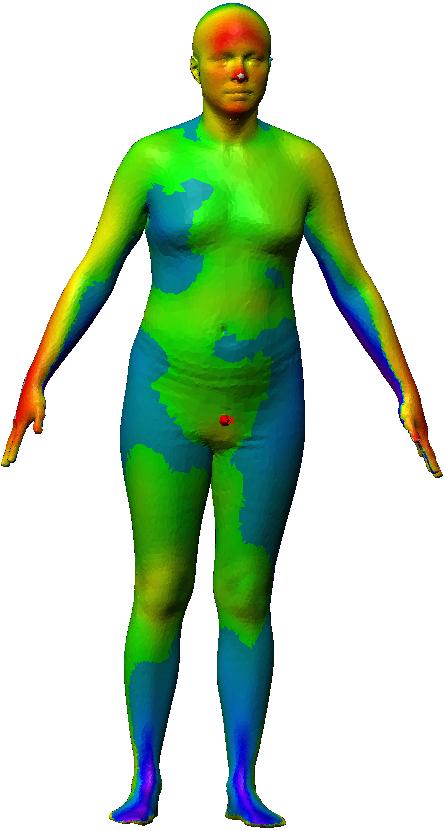} & 
\includegraphics[height = 6.0cm]{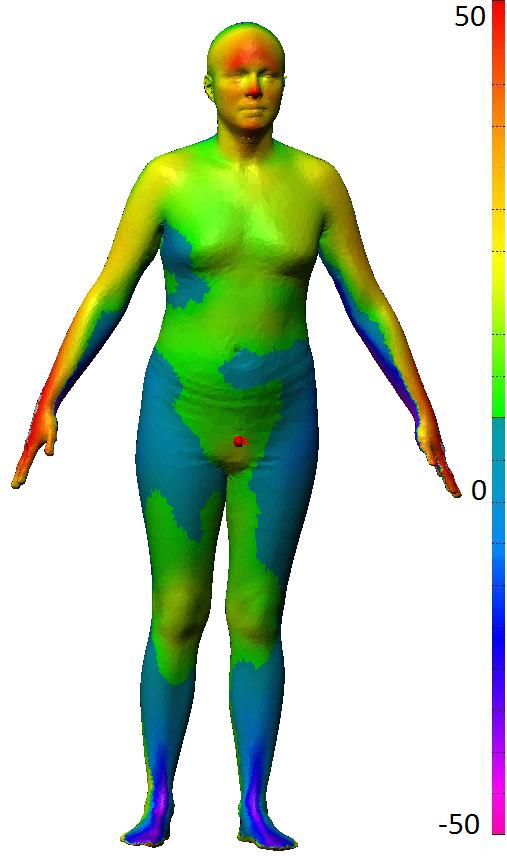} &
\includegraphics[height = 6.0cm]{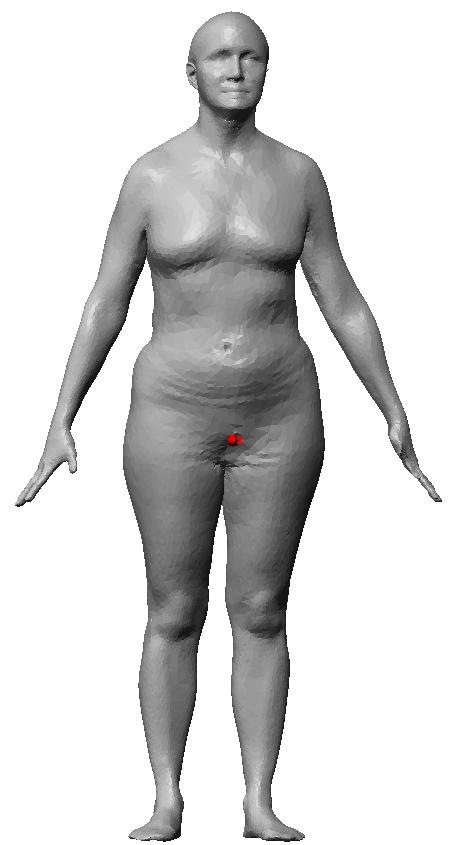} \\
\includegraphics[height = 6.0cm]{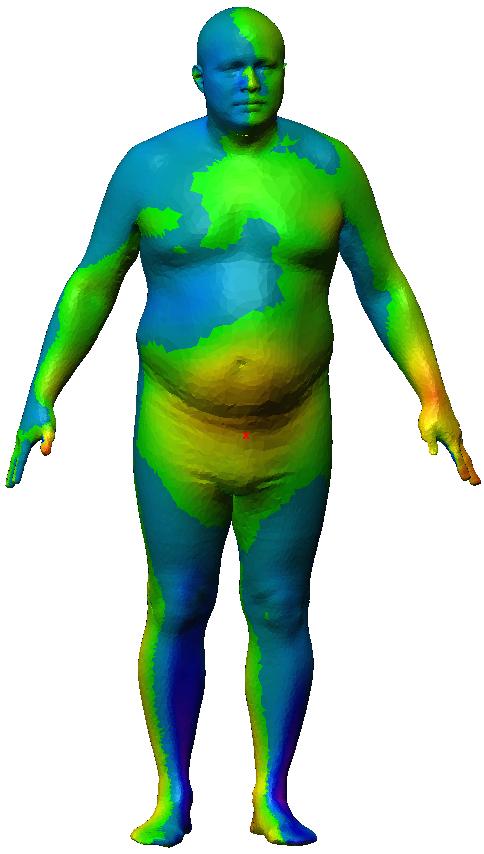} & 
\includegraphics[height = 6.0cm]{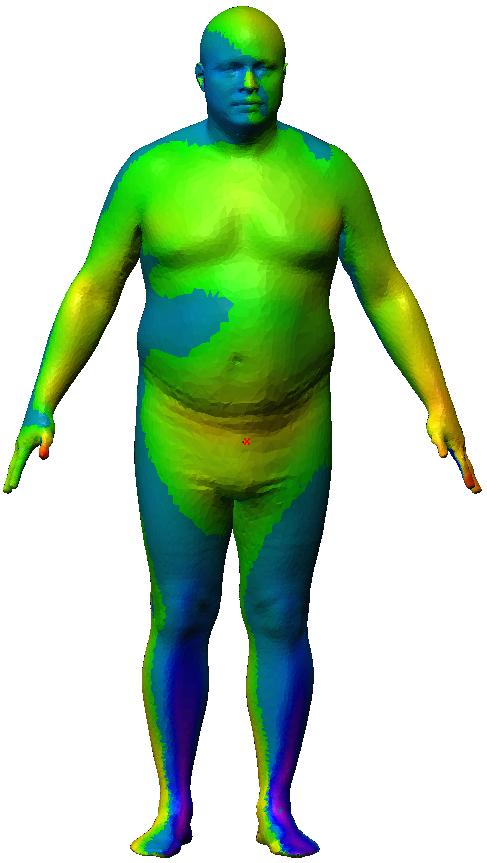} & 
\includegraphics[height = 6.0cm]{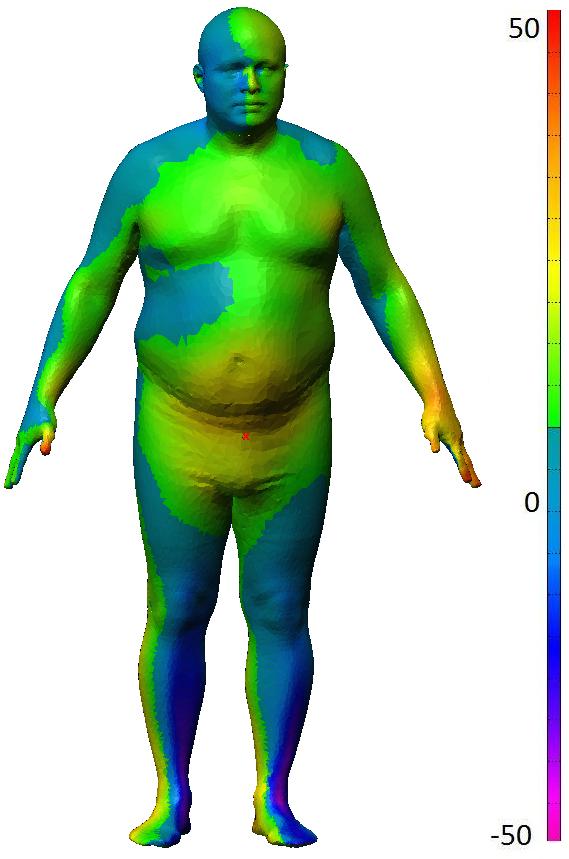} &
\includegraphics[height = 6.0cm]{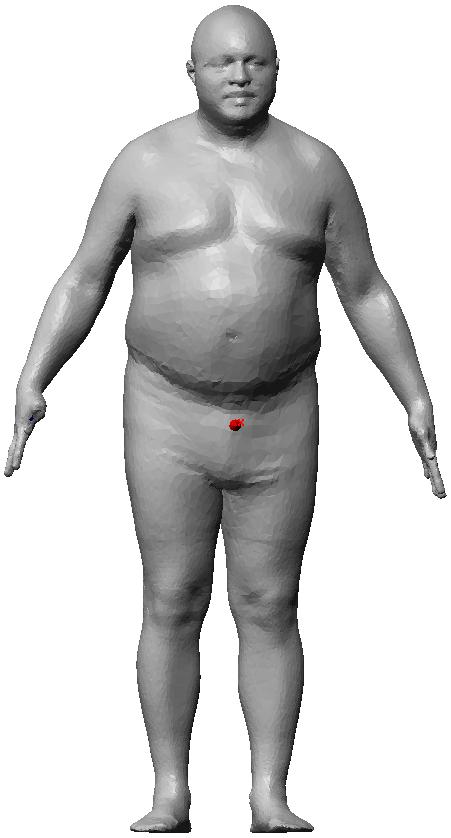} \\
(a) & (b) & (c) & (d) \\
\end{tabular}
\caption{\it{Predicted body shapes: (a) shows the result using our approach, (b) shows the result using SGPLVM, (c) shows the result using feature analysis, and (d) shows the ground truth. Each mesh is shown using a color coding corresponding to the signed distance from the parameterized body scan that was used to compute the measurements in $S_{real}$.}}
\label{true_bodies}
\end{figure*}

For the body shapes that aim to interpolate the measurements in $S_{real}$, Table~\ref{table:error} shows the errors of the Euclidean measurements and circumferences of the generated bodies. The error of the knee and arm circumferences is computed as the sum of the errors of the individual dimensions that contribute to the circumference. We compute the average and maximum errors in mm for each dimension. Note that our approach yields the lowest average errors for all dimensions and the lowest maximum errors for most dimensions. Figure~\ref{true_bodies} shows the predictions for three different subjects. Each mesh is shown using a color coding corresponding to the signed distance from the parameterized body scan that was used to compute the measurements in $S_{real}$. Note that SGPLVM yields predictions with the highest errors and also predictions that are visually most dissimilar from the ground truth. The predictions obtained using feature analysis and our approach are of similar quality. Note however that the predictions obtained using our approach are more accurate in localized areas of the body such as the waist area of the subject shown in the first row. This is due to the deformation of the mesh that aims to obtain the correct waist circumference.

\begin{table*}[htb]
\leavevmode
\centering
\footnotesize
\begin{tabular}{|l|r|r|r|r|r|r|}
\hline
 & \multicolumn{2}{|c|}{Feature Analysis} & \multicolumn{2}{|c|}{SGPLVM} & \multicolumn{2}{|c|}{Our Approach} \\
\hline
 & Avg. & Max. & Avg. & Max. & Avg. & Max. \\
\hline
Length Left Foot	&3.31	&10.01	&4.04	&11.47	&2.38e-4	&2.02e-3   \\
\hline
Length Left Lower Leg	&1.90	&6.50	&4.34	&11.30	&4.93e-4	&1.82e-3   \\
\hline
Length Left Upper Leg	&2.08	&6.79	&4.09	&12.07	&7.79e-4	&3.48e-3   \\
\hline
Length Right Upper Leg	&2.38	&7.67	&4.14	&25.71	&9.56e-4	&3.62e-3   \\
\hline
Length Right Lower Leg	&1.99	&5.48	&5.00	&18.91	&5.57e-4	&2.49e-3   \\
\hline
Length Right Foot	&4.37	&14.33	&4.84	&16.70	&2.09e-4	&2.28e-3   \\
\hline
Length Upper Body	&1.71	&4.90	&6.93	&27.67	&2.99e-4	&9.91e-4   \\
\hline
Right Shoulder-Neck Distance	&1.75	&6.99	&3.06	&9.946	&1.71e-3	&6.65e-3   \\
\hline
Length Right Upper Arm	&2.21	&7.78	&3.73	&10.98	&6.11e-4	&1.74e-3   \\
\hline
Length Right Lower Arm	&3.17	&8.89	&4.36	&16.95	&4.42e-4	&1.79e-3   \\
\hline
Left Shoulder-Neck Distance	&2.37	&9.85	&2.82	&12.77	&2.12e-3	&8.50e-3   \\
\hline
Length Left Upper Arm	&4.24	&12.36	&5.16	&19.17	&7.16e-4	&2.2e-3   \\
\hline
Length Left Lower Arm	&2.46	&8.35	&4.89	&14.14	&4.79e-4	&2.21e-3   \\
\hline
Head-Neck Distance	&4.03	&12.58	&4.47	&12.57	&1.14e-3	&5.38e-3   \\
\hline
Circumference Right Knee	&6.61	&14.12	&6.78	&12.82	&5.48	&15.36   \\
\hline
Circumference Left Knee	&6.97	&24.40	&7.12	&18.40	&6.02	&24.44   \\
\hline
Circumference Right Arm	&7.98	&18.35	&8.55	&22.62	&4.62	&11.17   \\
\hline
Circumference Left Arm	&7.38	&16.66	&8.42	&18.13	&5.07	&16.36   \\
\hline
Hip Circumference	&10.20	&26.83	&18.97	&110.30	&5.29	&40.01   \\
\hline
Waist Circumference	&8.99	&32.13	&15.51	&45.19	&1.24	&6.11   \\
\hline
Chest Circumference	&8.32	&22.50	&15.87	&89.53	&1.29	&4.08   \\
\hline
Head Circumference	&10.56	&27.99	&11.67	&28.75	&8.414	&26.76   \\
\hline
\end{tabular}
\caption{\it{Average and maximum errors (in mm) of measurements over 50 samples of $S_{real}$.}}
\label{table:error}
\end{table*}

\subsubsection*{Experiments Using Real Measurements}

This section presents an experiment that aims to predict the body shape from a set of measurements that were measured on real subjects using a tape measure. For this experiment, we consider the measurements of six volunteers who measured each other. Since the measurements of interest are difficult to measure for non-experts, we expect errors in the input data. The parameters are therefore set to $l=3$ to restrict the shapes to the space of the training data and to $\lambda=1.0$ to avoid fine fitting to inaccurate measurements. Furthermore, we found that the measurements of the knee and arm circumferences were unreliable in practice, so we did not use these measurements to generate the reconstructions. 

\begin{figure*}[htb]
\centering
\begin{tabular}{c c c c c c}
\includegraphics[height = 4.0cm]{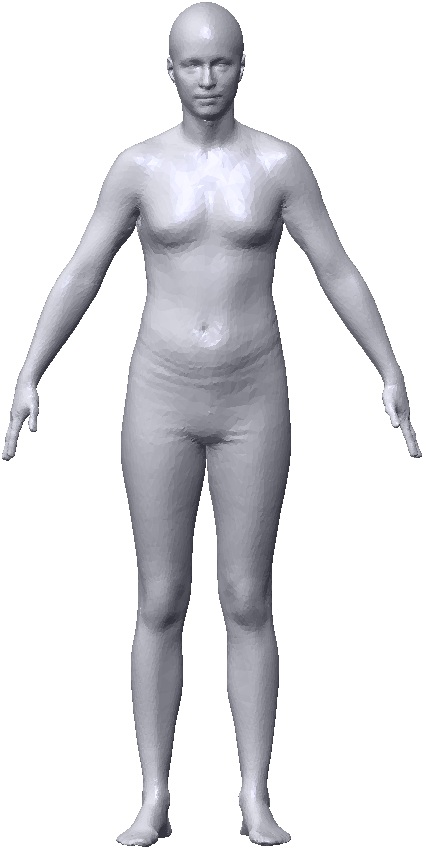} &
\includegraphics[height = 4.0cm]{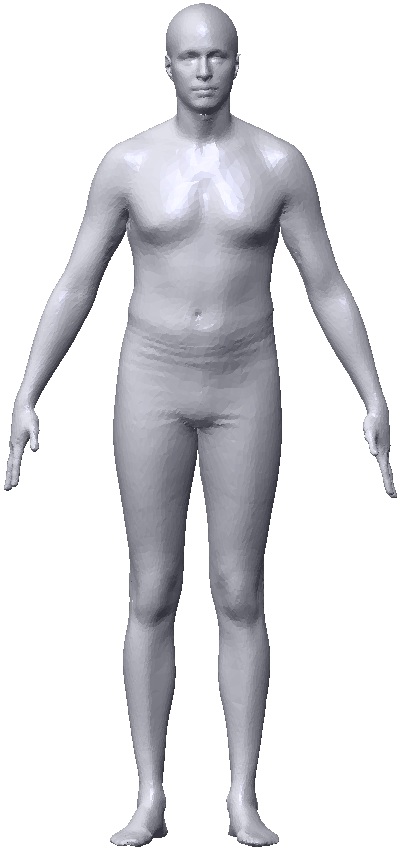} &
\includegraphics[height = 4.0cm]{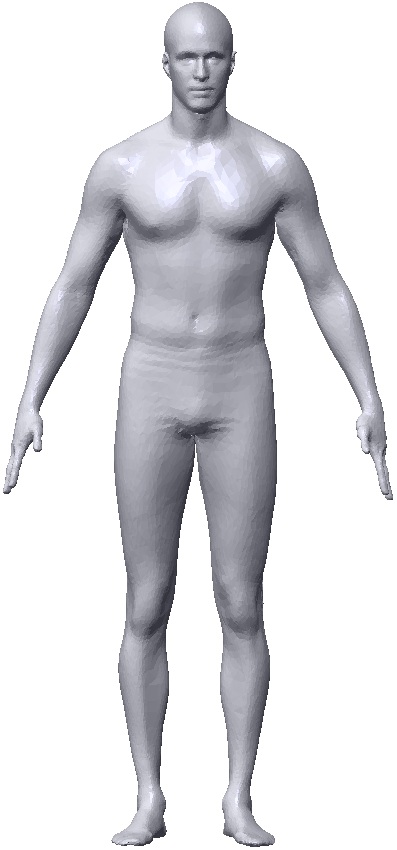} &
\includegraphics[height = 4.0cm]{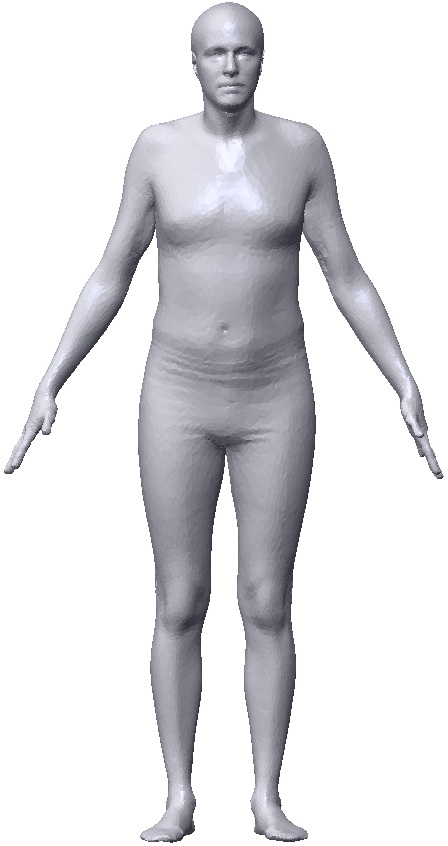} &
\includegraphics[height = 4.0cm]{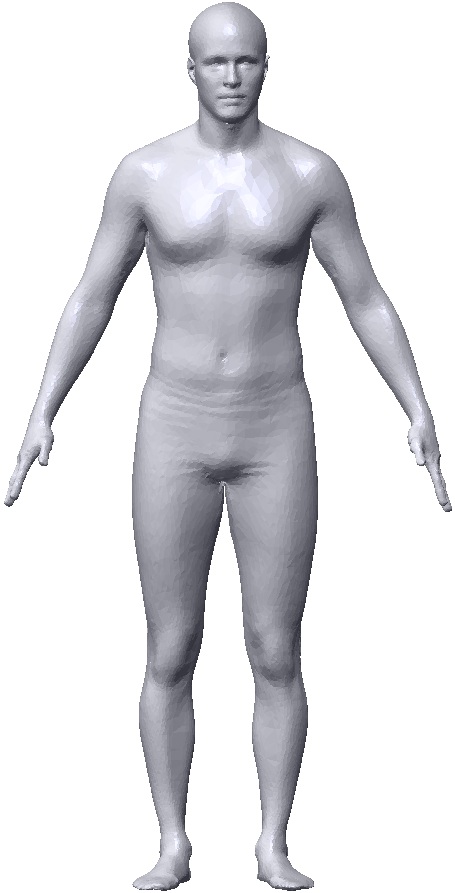} &
\includegraphics[height = 4.0cm]{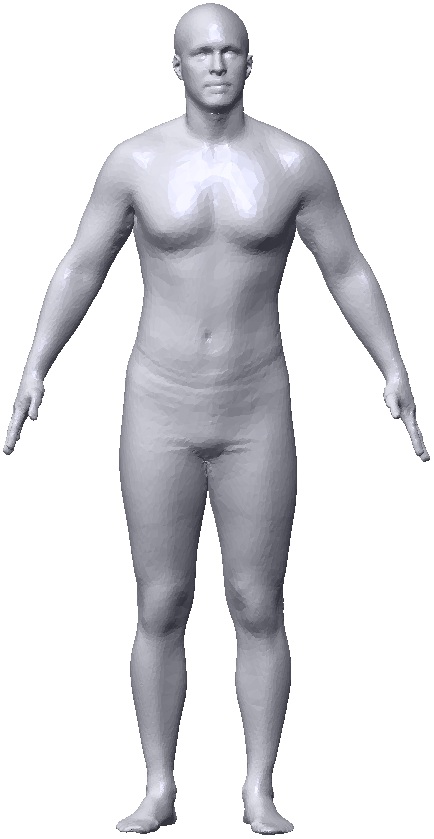} \\
\includegraphics[height = 4.0cm]{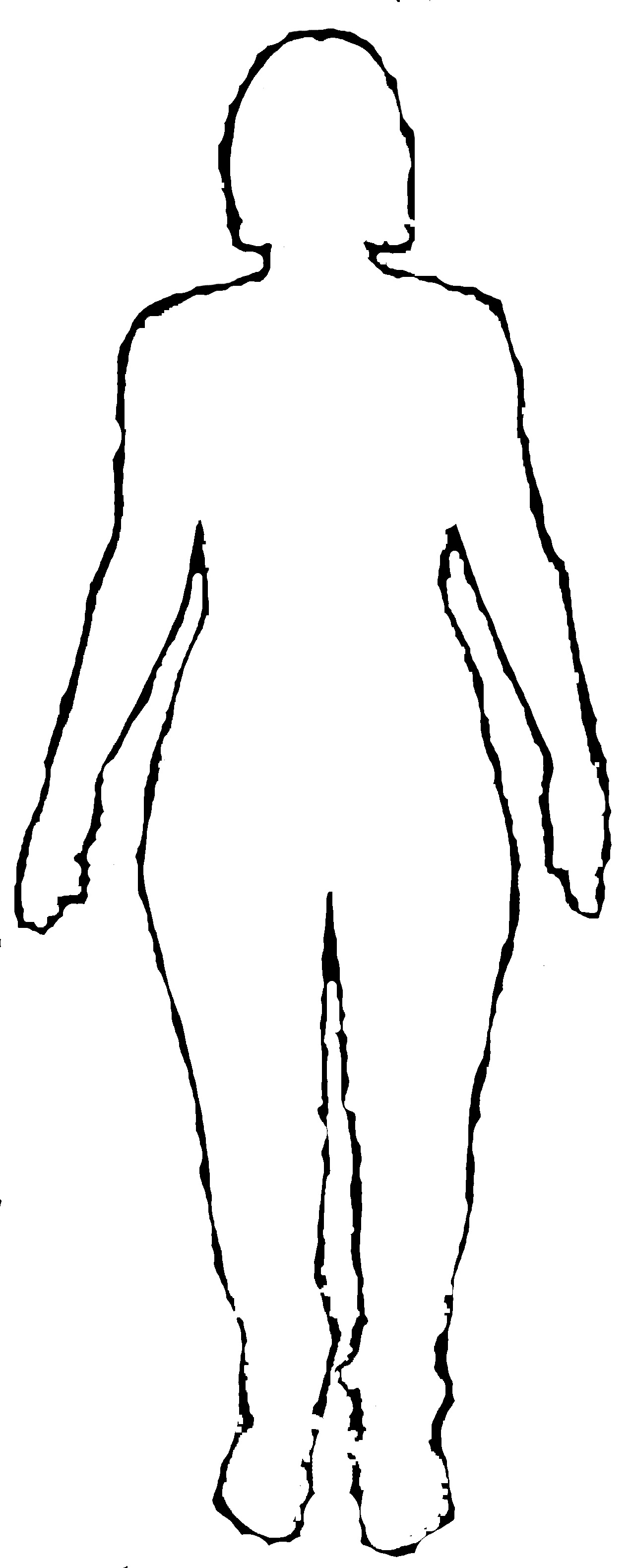} &
\includegraphics[height = 4.0cm]{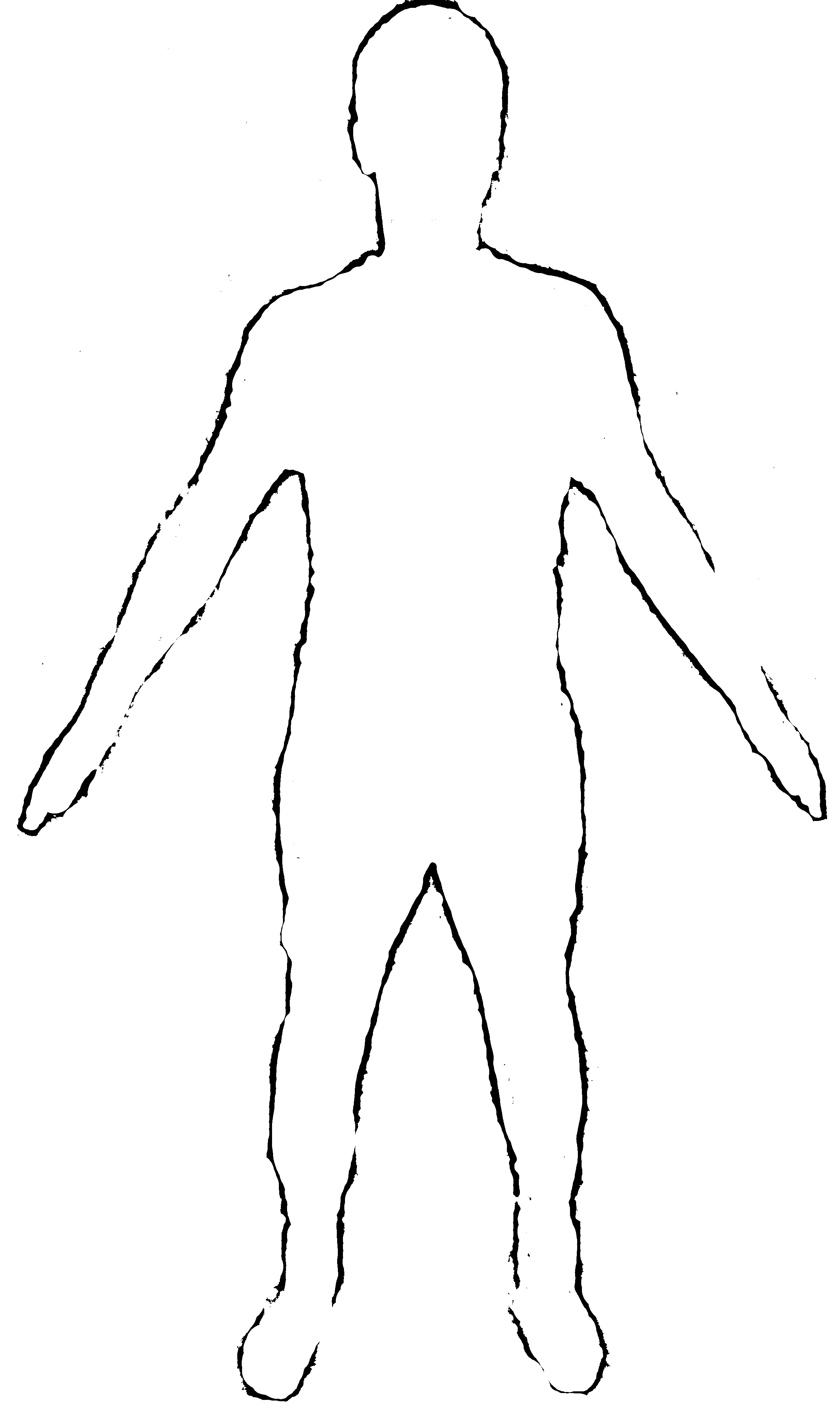} &
\includegraphics[height = 4.0cm]{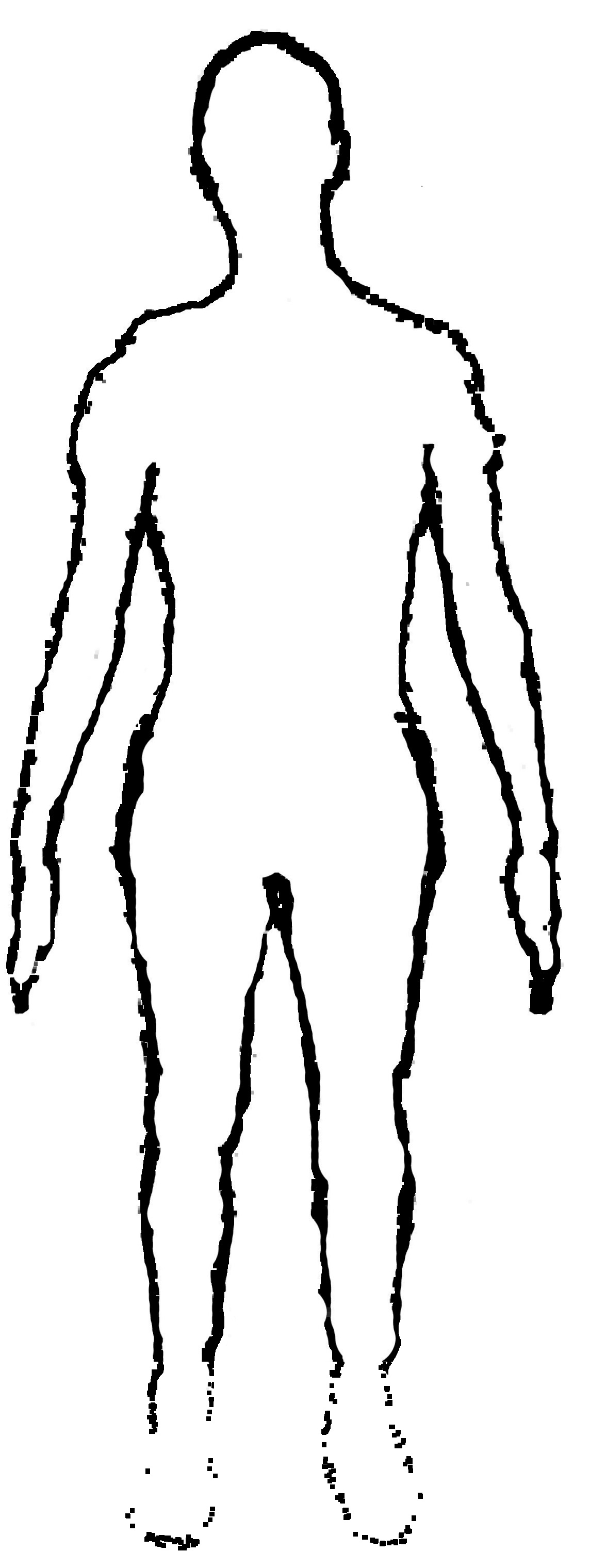} &
\includegraphics[height = 4.0cm]{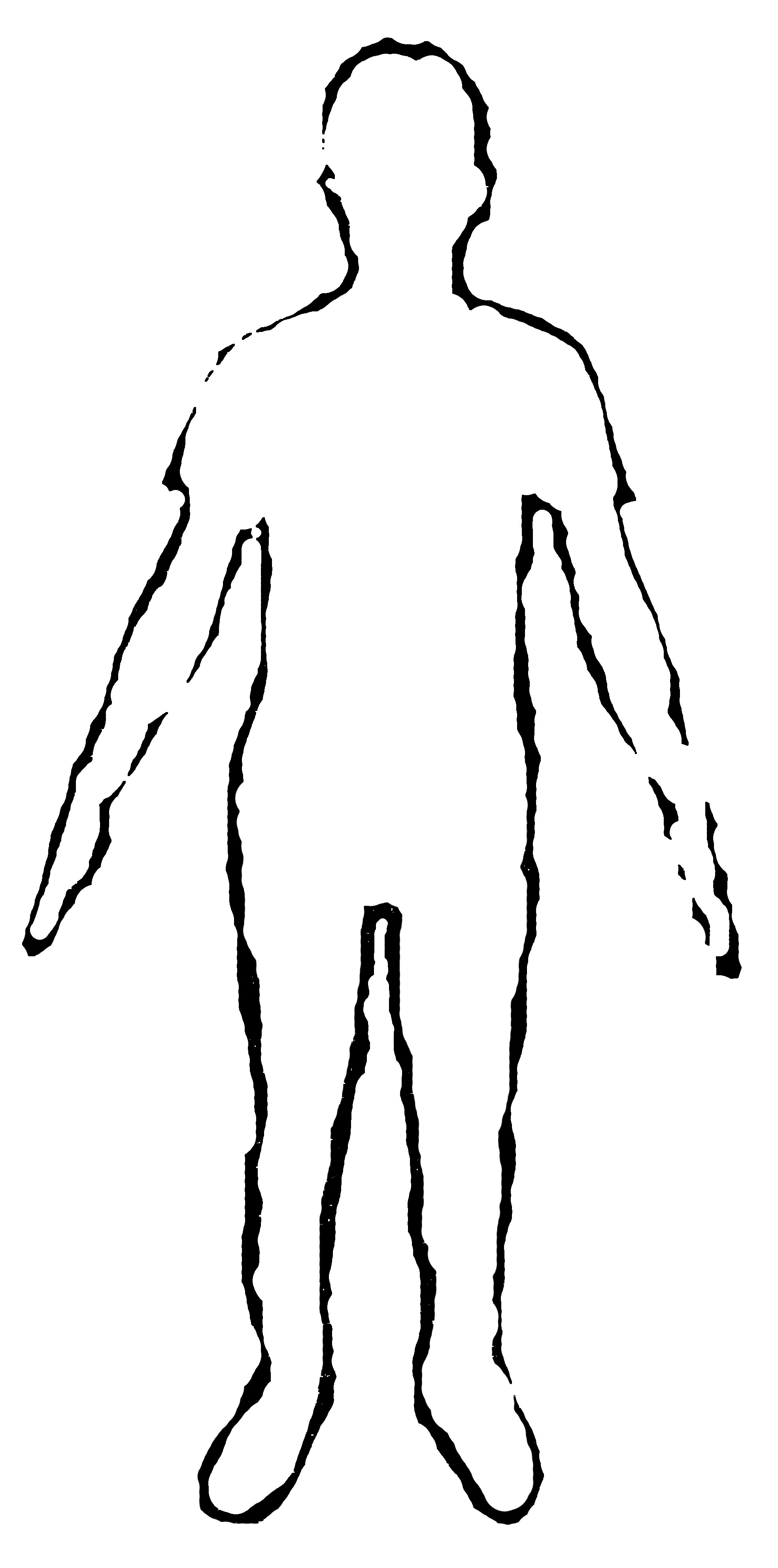} &
\includegraphics[height = 4.0cm]{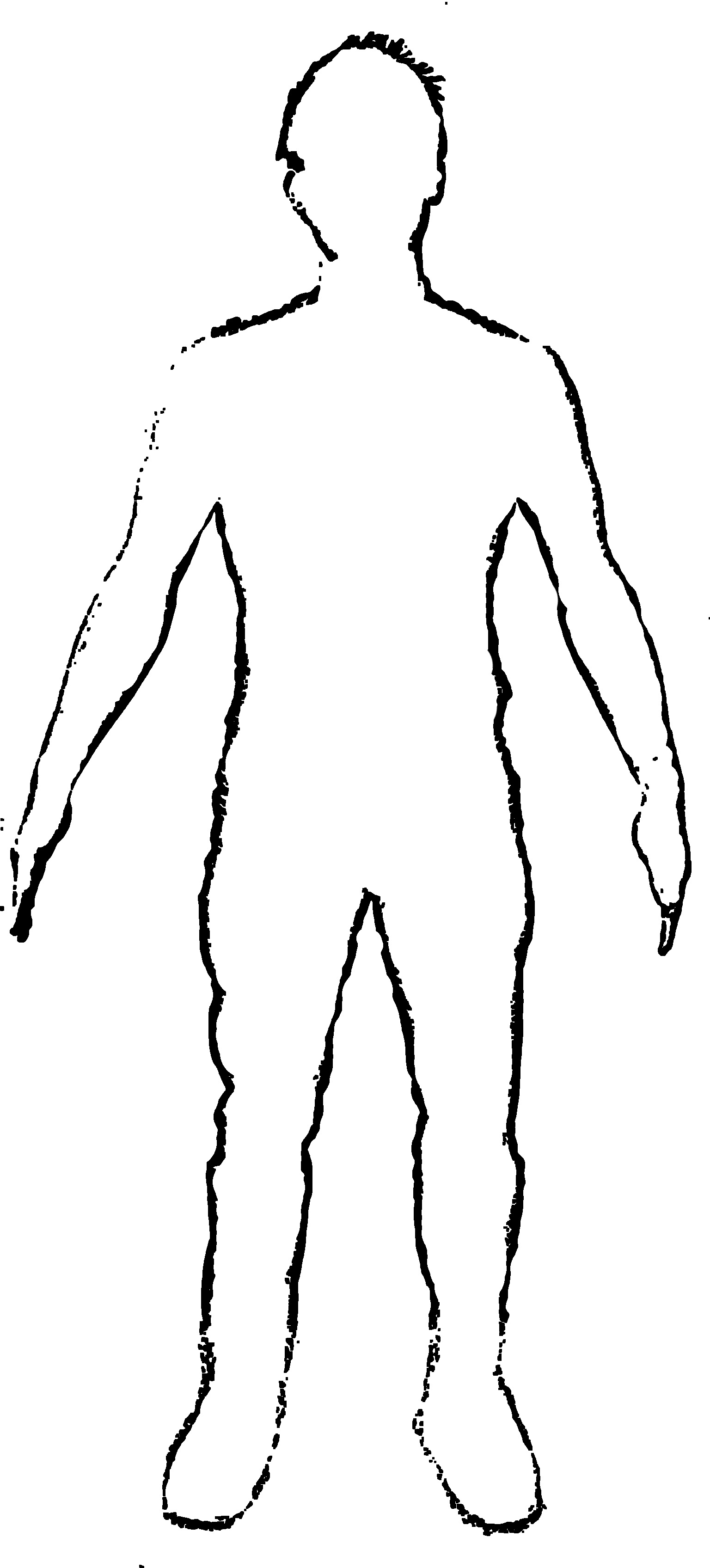} &
\includegraphics[height = 4.0cm]{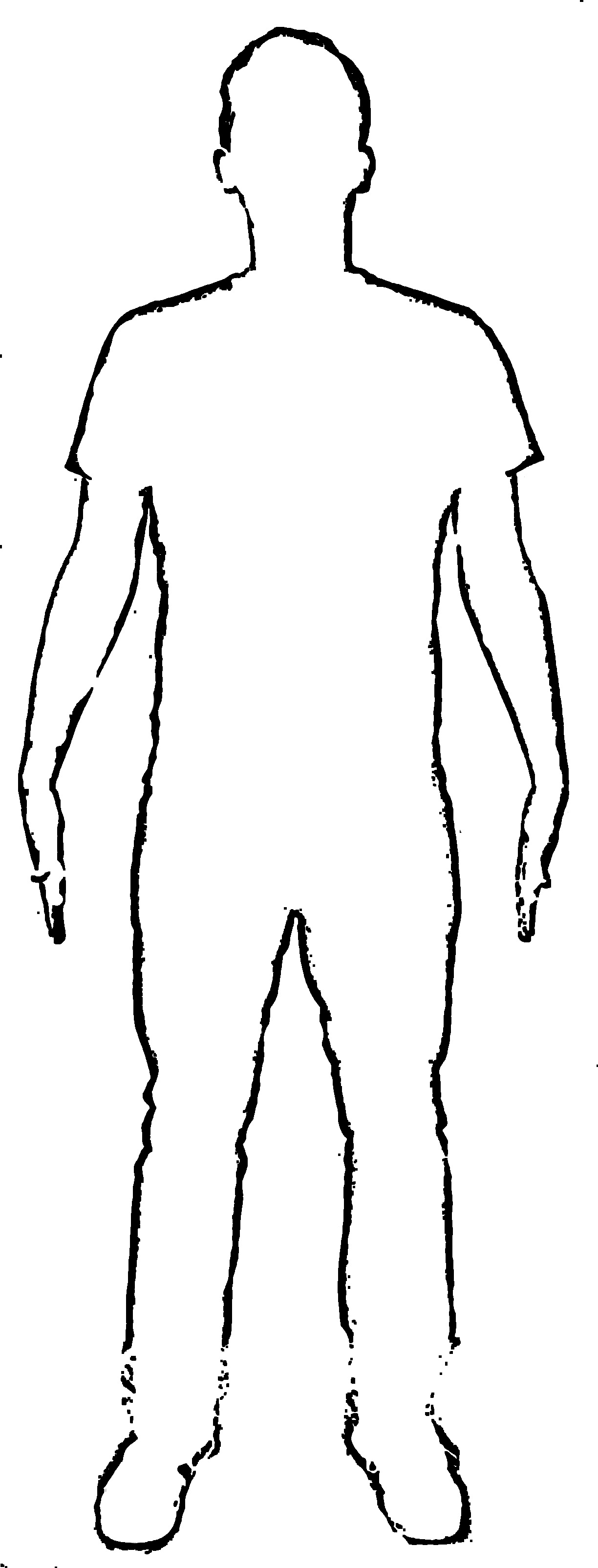} \\
\includegraphics[height = 4.0cm]{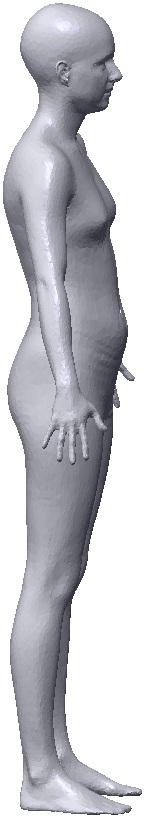} &
\includegraphics[height = 4.0cm]{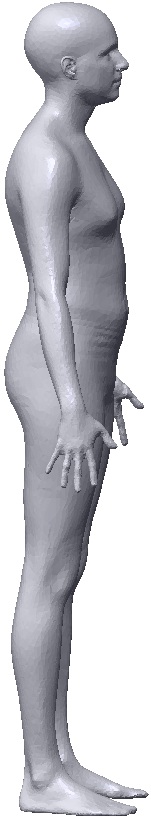} &
\includegraphics[height = 4.0cm]{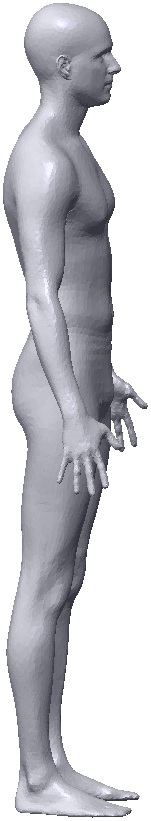} &
\includegraphics[height = 4.0cm]{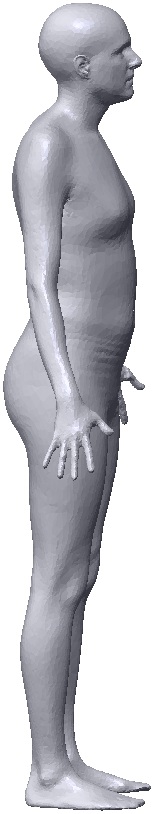} &
\includegraphics[height = 4.0cm]{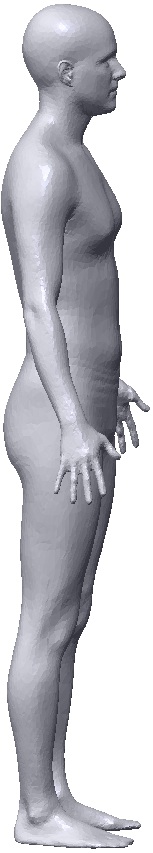} &
\includegraphics[height = 4.0cm]{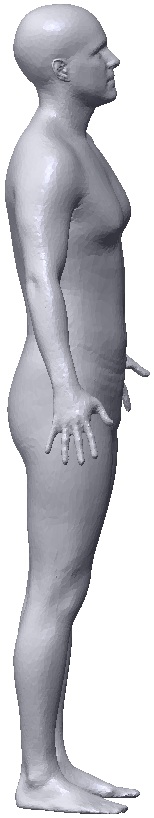} \\
\includegraphics[height = 4.0cm]{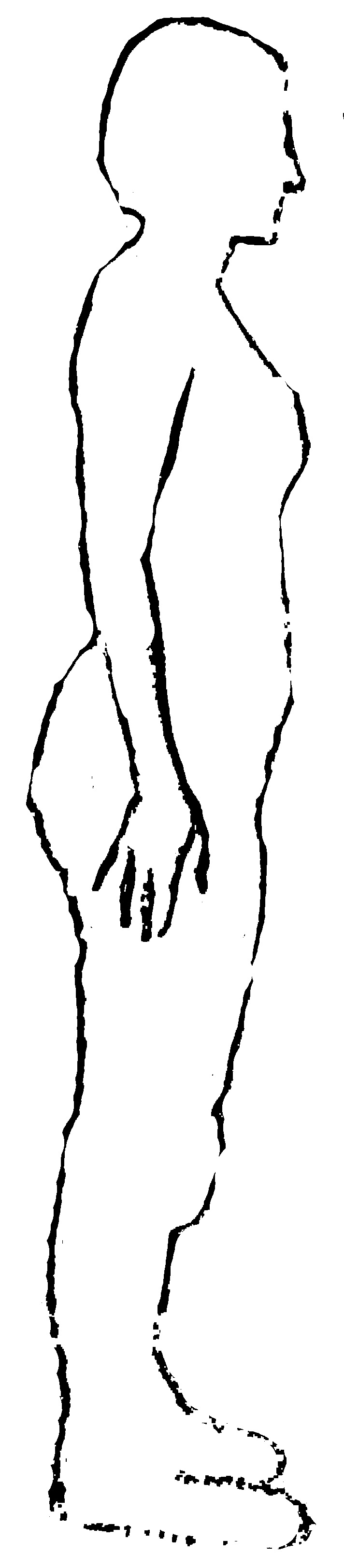} &
\includegraphics[height = 4.0cm]{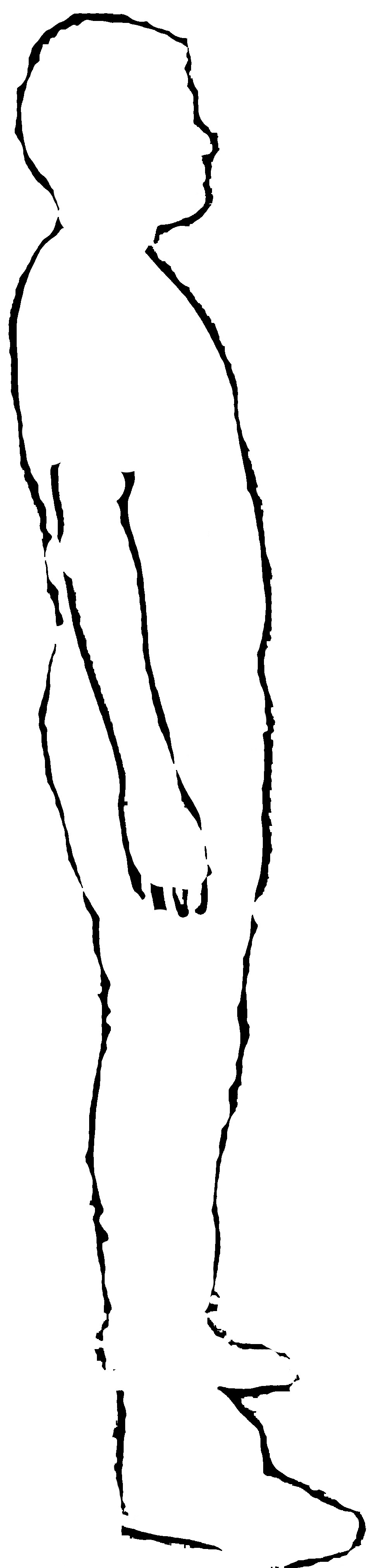} &
\includegraphics[height = 4.0cm]{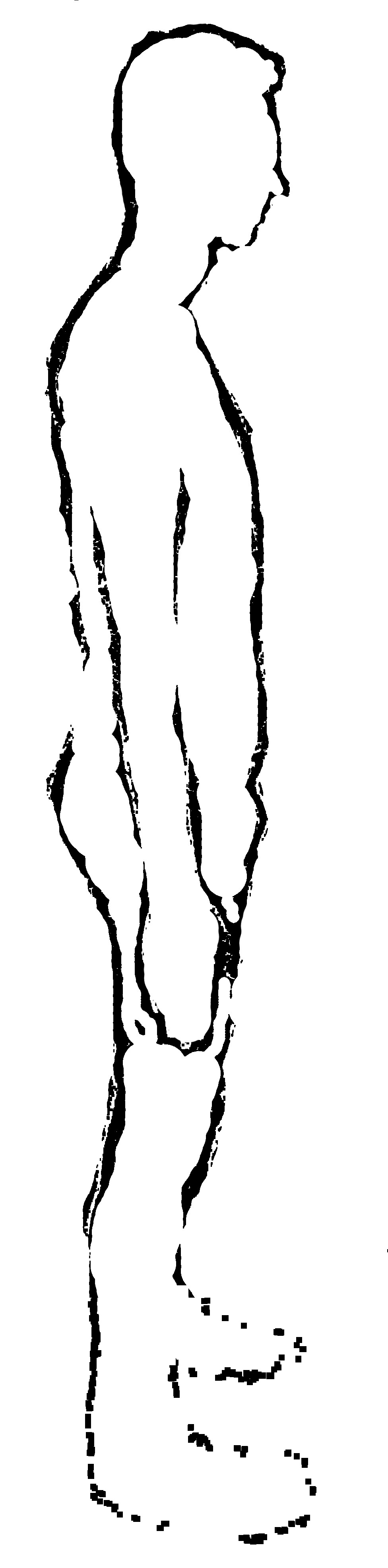} & 
\includegraphics[height = 4.0cm]{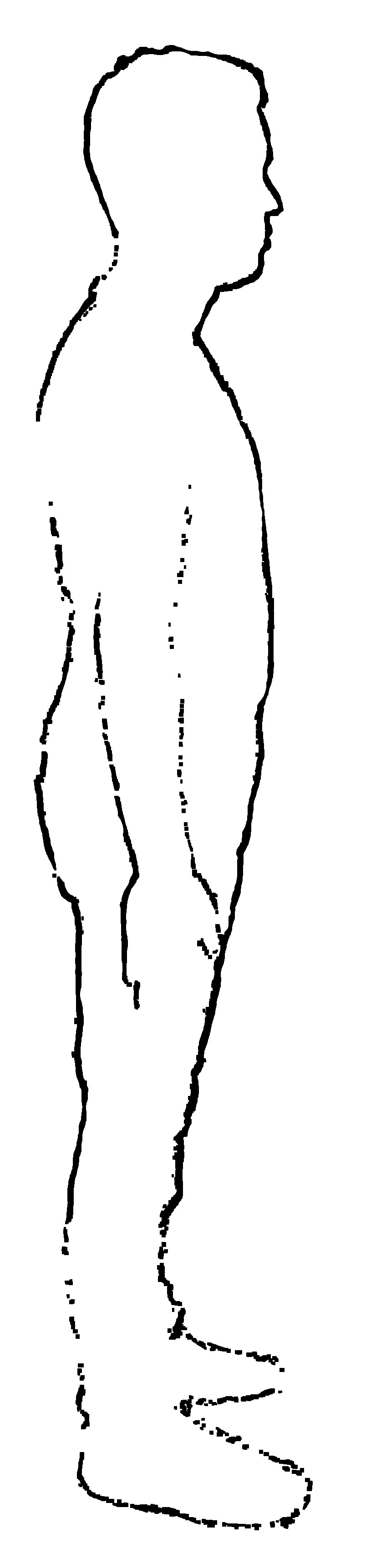} & 
\includegraphics[height = 4.0cm]{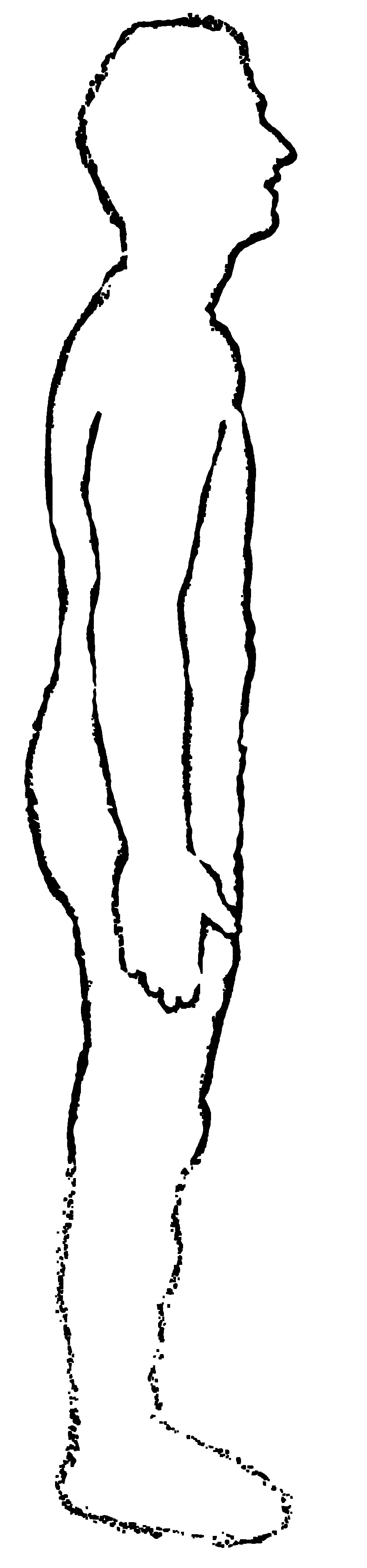} & 
\includegraphics[height = 4.0cm]{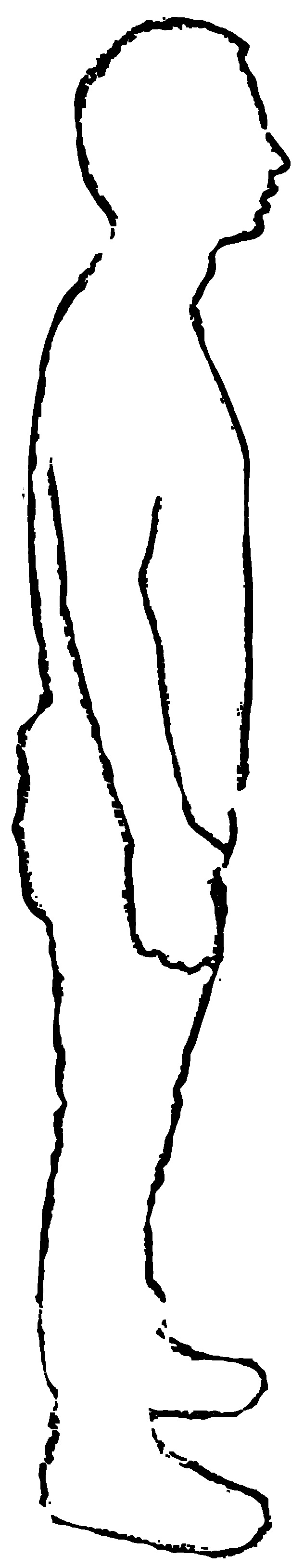} \\
\end{tabular}
\caption{\it{Predicted body shapes from real measurements acquired by non-experts. Each column shows for one subject the reconstructed 3D model and two silhouette images extracted from photographs of the clothed subject.}}
\label{real_measurements}
\end{figure*}

Figure~\ref{real_measurements} shows the results. Each column shows for one subject screen shots of the predicted body shape and silhouette images extracted from photographs of the clothed subject taken from similar viewpoints. Note that since the silhouettes are taken from clothed subjects, the silhouettes only approximate the silhouettes of the true body shapes (in fact, the silhouettes in the images should be wider than the true silhouettes). Furthermore, the 3D models and the silhouettes may show the subjects in slightly different postures. We can observe that even with the limited accuracy of both the measurements and the silhouettes, we obtain 3D body shape predictions that are visually consistent with the silhouette images. 

Table~\ref{table:error_real} shows the errors of the measurements. Note that the errors are overall consistent with the errors reported for digitally measured distances in Table~\ref{table:error}, albeit significantly higher. There are two main reasons for the higher errors. First, the measurements were conducted by non-experts. Gordon et al.~\cite{gordon_etal} reported that measurements can be repeated with an accuracy of about 1cm when conducted by trained experts. In our scenario, we expect significantly higher measurement errors. Second, some of the measurements do not pass through salient surface points and are difficult to measure. Hence, the manual measurements and the digital measurements (i.e. the measurements computed from the 3D model by the algorithm) may not measure exactly the same distances. In spite of these problems, all measurement errors are lower than 6cm. This accuracy may suffice for applications, such as virtual games or virtual try-on.

\begin{table*}[htb]
\leavevmode
\centering
\footnotesize
\begin{tabular}{|l|r|r|r|r|r|r|}
\hline
 & Subject 1 & Subject 2 & Subject 3 & Subject 4 & Subject 5 & Subject 6 \\
\hline
Length Left Foot	& 2.21 & 8.26 & 4.31 & 18.36 & 6.42 & 4.80 \\
\hline
Length Left Lower Leg	& 8.38 & 0.67 & 4.60 & 3.11 & 6.37 & 2.58 \\
\hline
Length Left Upper Leg	& 1.27 & 3.36 & 8.47 & 1.14 & 8.97 & 14.98 \\
\hline
Length Right Upper Leg	& 6.51 & 0.86 & 12.02 & 4.21 & 13.05 & 7.52 \\
\hline
Length Right Lower Leg	& 8.19 & 3.34 & 7.57 & 2.64 & 9.84 & 6.74 \\
\hline
Length Right Foot	& 7.91 & 0.64 & 0.23 & 10.95 & 1.56 & 6.28 \\
\hline
Length Upper Body	& 1.02 & 0.50 & 1.13 & 5.76 & 0.84 & 6.35  \\
\hline
Right Shoulder-Neck Distance	& 4.68 & 2.45 & 3.22 & 14.81 & 4.58 & 5.12 \\
\hline
Length Right Upper Arm	& 14.65 & 2.41 & 5.19 & 13.92 & 0.61 & 5.32 \\
\hline
Length Right Lower Arm	& 3.84 & 0.47 & 1.21 & 4.50 & 0.18 & 13.10  \\
\hline
Left Shoulder-Neck Distance	& 4.13 & 5.62 & 5.72 & 6.87 & 3.23 & 11.89  \\
\hline
Length Left Upper Arm	& 12.93 & 0.87 & 5.74 & 15.89 & 2.02 & 15.22 \\
\hline
Length Left Lower Arm	& 5.53 & 4.67 & 2.97 & 5.79 & 1.53 & 4.61  \\
\hline
Head-Neck Distance	&  12.24 & 5.98 & 11.12 & 3.70 & 10.15 & 30.29 \\
\hline
Hip Circumference	& 29.86 & 7.56 & 24.99 & 17.71 & 28.83 & 27.20  \\
\hline
Waist Circumference	& 3.53 & 12.75 & 2.89 & 24.34 & 7.94 & 15.56  \\
\hline
Chest Circumference	& 53.85 & 4.02 & 13.64 & 14.23 & 15.07 & 25.94  \\
\hline
Head Circumference	&  39.66 & 6.91 & 40.02 & 3.57 & 30.48 & 19.71 \\
\hline
\end{tabular}
\caption{\it{Errors (in mm) of measurements for each of the six subjects shown in Figure~\ref{real_measurements}.}}
\label{table:error_real}
\end{table*}

\subsubsection*{Experiments With a Small Training Set} 

\begin{figure*}[htb]
\centering
\includegraphics[width = \textwidth]{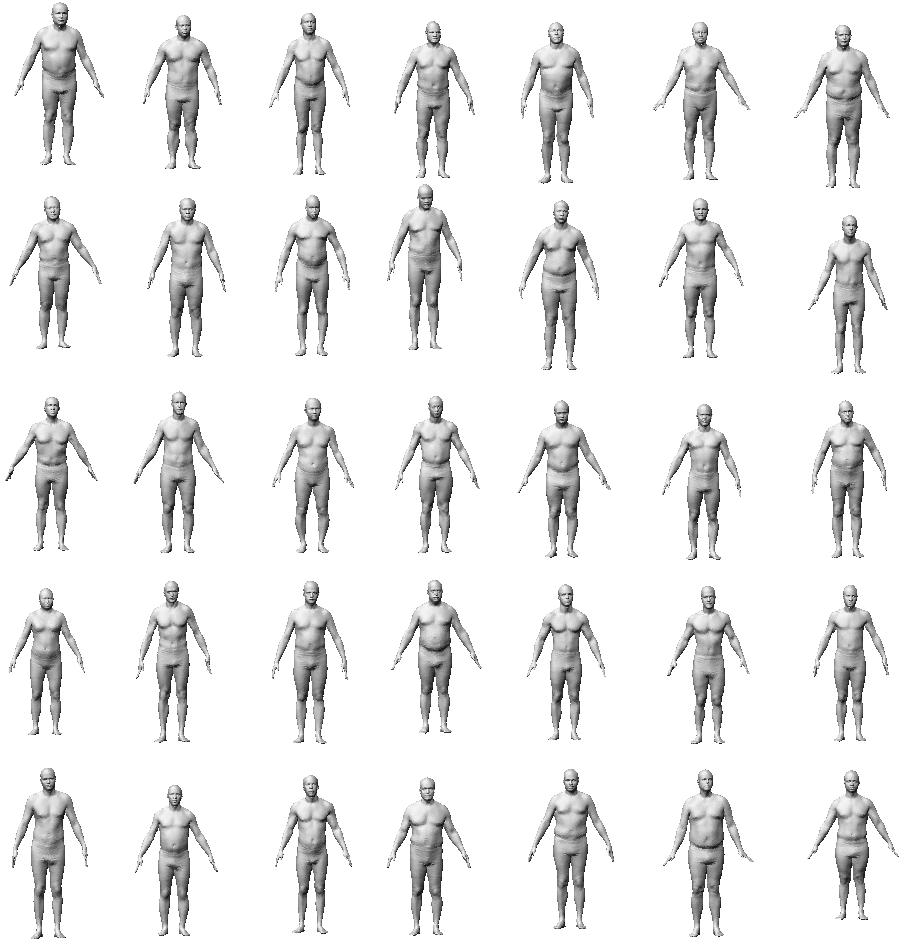}
\caption{\it{35 bodies used for training.}}
\label{typical_male_training}
\end{figure*}

Finally, we demonstrate the ability of our approach to predict shapes that are not covered well by the training data. Recall that this is relevant in practice because predicting human shapes based a small training sample avoids the complicated and expensive acquisition of large anthropometric databases. We use as training data a set of 35 of the bodies of the CAESAR database. All of the bodies in the training set are male bodies with typical body shape, see Figure~\ref{typical_male_training}. 

We choose five additional male bodies with body shape variations not covered by the training data and we use each of these models to digitally measure the 34 relevant measurements. We then use these measurements for prediction. For this experiment, we set $l=3$ since the shapes are not represented well by the training data. Furthermore, we set $s=10$ to increase the accuracy of the prediction. 

\begin{figure*}[htb]
\centering
\includegraphics[width = \textwidth]{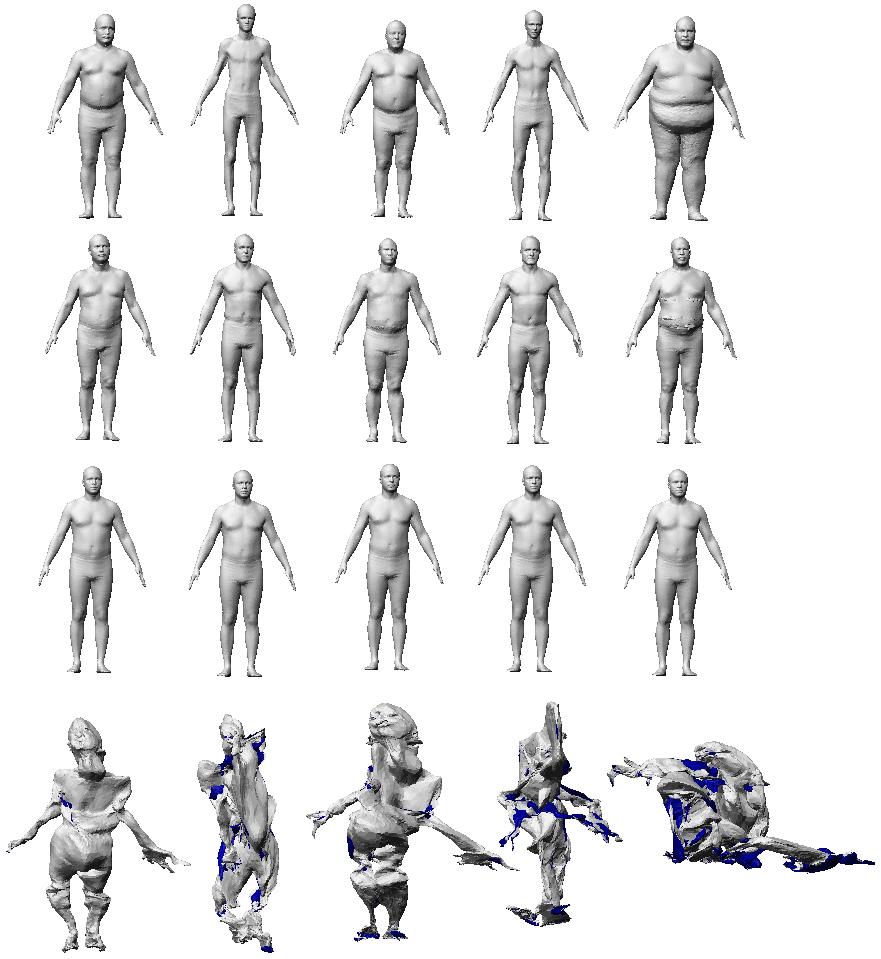}
\caption{\it{Predicted body shapes based on the training data set shown in Figure~\ref{typical_male_training}. The first row shows the ground truth, the second row shows the result using our approach, the third row shows the result using SGPLVM, and the last row shows the result using feature analysis.}}
\label{atypical_prediction}
\end{figure*}

Figure~\ref{atypical_prediction} shows the results. The first row shows the bodies of the CAESAR database used for prediction, the second row shows the result using our approach, the third row shows the result using SGPLVM, and the last row shows the result using feature analysis. Note that our approach predicts body shapes that are similar to the ground truth for most shapes although a limited training set was used. For the body in the last column, the result using our approach is not as large as the ground truth. The reason is that the input body shape is far from the training data. The results using SGPLVM are all close to the mean shape and far from the ground truth. The results using feature analysis are outside of the shape space of human body shapes. This shows that our approach best handles the case where the predicted shapes have variations that are outside of the shape space spanned by the training data.

\subsection{Summary}

We conducted experiments based on human face and body shapes using both real and synthetic data. We conducted an extensive evaluation by estimating 600 face shapes and over 650 body shapes. In summary, we demonstrated experimentally that our method has the following properties.

\begin{itemize}
\item Our method estimates realistic body or face shapes that are outside of the learned shape space.
\item Our method yields results that are more accurate than SGPLVM, and that, unlike feature analysis, always stay within the shape space of human body or face shapes.
\item Our method accurately predicts local shape variations not present in the training set (which accounts for accurate predictions of the waist shapes in Figure~\ref{true_bodies}).
\item Our method yields visually pleasing results even when a small database is used for training. This is important in practice because the acquisition of a large database for training is costly.
\item Even for inaccurate input measurements acquired by non-experts, our method is able to find visually consistent results.
\item Our method is computationally efficient. The running time of the optimization steps is linear in the number of vertices. To predict one human body shape, our (non optimized) implementation takes about 35 seconds on a standard PC with a dual core CPU and 8GB of RAM.
\end{itemize}

\section{Conclusion}

We presented a novel approach to estimate human body and face shapes based on measurements. We demonstrated that the approach can produce a large variety of likely shapes based on a relatively small training set and a small number of measurements. We showed experimentally that, unlike feature analysis, the approach always produces shapes that are inside the space of human shapes. Furthermore, the approach yields higher accuracy than SGPLVM. That is, unlike previous approaches, our approach predicts shape variations not present in the training data while maintaining a realistic human body shape. This allows to generate accurate body shape estimates based on simple modalities without the need to acquire a large-scale 3D database that represents the target population.

\section*{Acknowledgments}
We thank the volunteers for participating in the experiment. Furthermore, we thank Pengcheng Xi for helpful discussions and for providing us with the training data, Neil Lawrence for providing us with the SGPLVM code, and the anonymous reviewers for helpful comments. This work has partially been funded by the Cluster of Excellence \textit{Multimodal Computing and Interaction} within the Excellence Initiative of the German Federal Government.

\small{

}


\begin{thebibliography}{10}

\bibitem{allen_curless_popovic_03_parametrization_body_shape}
B.~Allen, B.~Curless, and Z.~Popovi\'{c}.
\newblock The space of human body shapes: reconstruction and parameterization
  from range scans.
\newblock {\em ACM Transactions on Graphics}, 22(3):587--594, 2003.
\newblock Proceedings of SIGGRAPH.

\bibitem{allen_curless_popovic_04}
B.~Allen, B.~Curless, and Z.~Popovi\'{c}.
\newblock Exploring the space of human body shapes: Data-driven synthesis under
  anthropometric control.
\newblock In {\em SAE Symposium on Digital Human Modeling for Design and
  Engineering}, 2004.

\bibitem{anguelov_srinivasan_koller_thrun_rodgers_05_shapecomp}
D.~Anguelov, P.~Srinivasan, D.~Koller, S.~Thrun, J.~Rodgers, and J.~Davis.
\newblock Scape: shape completion and animation of people.
\newblock {\em ACM Transactions on Graphics}, 24(3):408--416, 2005.
\newblock Proceedings of SIGGRAPH.

\bibitem{Baek2010}
S.-Y. Baek and K.~Lee.
\newblock Parametric human body shape modeling framework for human-centered
  product design.
\newblock {\em Computer-Aided Design}, To appear.

\bibitem{Barber96thequickhull}
C.~B. Barber, D.~P. Dobkin, and H.~Huhdanpaa.
\newblock The quickhull algorithm for convex hulls.
\newblock {\em ACM Transactions on Mathematical Software}, 22(4):469--483,
  1996.

\bibitem{blanz_vetter_99}
V.~Blanz and T.~Vetter.
\newblock A morphable model for the synthesis of 3d faces.
\newblock In {\em Proceedings of SIGGRAPH}, pages 187--194, 1999.

\bibitem{chen_cipolla_SPGLVM_reconstruction}
Y.~Chen and R.~Cipolla.
\newblock Learning shape priors for single view reconstruction.
\newblock In {\em Workshop on 3D Imaging and Modelling}, 2009.

\bibitem{chu_etal_2010}
C.-H. Chu, Y.-T. Tsai, C.~C. Wang, and T.-H. Kwok.
\newblock Exemplar-based statistical model for semantic parametric design of
  human body.
\newblock {\em Computers in Industry}, 61(6):541--549, 2010.

\bibitem{decarlo_etal_98}
D.~DeCarlo, D.~Metaxas, and M.~Stone.
\newblock An anthropometric face model using variational techniques.
\newblock In {\em Proceedings of SIGGRAPH}, pages 67--74, 1998.

\bibitem{Dijkstra1959}
E.~W. Dijkstra.
\newblock A note on two problems in connexion with graphs.
\newblock {\em Numerische Mathematik}, 1:269--271, 1959.

\bibitem{dryden_mardia_shape_analysis}
I.~Dryden and K.~Mardia.
\newblock {\em Statistical Shape Analysis}.
\newblock Wiley, 2002.

\bibitem{ek_torr_lawrence_07}
C.~H. Ek, P.~H.~S. Torr, and N.~D. Lawrence.
\newblock Gaussian process latent variable models for human pose estimation.
\newblock In {\em Workshop on Machine Learning for Multimodal Interaction},
  pages 132--143, 2007.

\bibitem{gordon_etal}
C.~C. Gordon, T.~Churchill, C.~E. Clauser, B.~Bradtmiller, J.~T. McConville,
  I.~Tebberets, and R.~A. Walker.
\newblock Anthropometric survey of {U.S.} army personnel: Methods and summary
  statistics 1988.
\newblock Technical report, U.S. Army Natick Research, Development, and
  Engineering Center, 1989.

\bibitem{guan_etal}
P.~Guan, A.~Weiss, A.~O. Balan, and M.~J. Black.
\newblock Estimating human shape and pose from a single image.
\newblock In {\em International Conference on Computer Vision}, 2009.

\bibitem{hasler_ackermann_etal_10}
N.~Hasler, H.~Ackermann, B.~Rosenhahn, T.~Thorm\"{a}hlen, and H.-P. Seidel.
\newblock Multilinear pose and body shape estimation of dressed subjects from
  image sets.
\newblock In {\em Conference on Computer Vision and Pattern Recognition}, 2010.

\bibitem{HasStoSunRosSei09}
N.~Hasler, C.~Stoll, M.~Sunkel, B.~Rosenhahn, and H.-P. Seidel.
\newblock A statistical model of human pose and body shape.
\newblock In P.~Dutr\'{e} and M.~Stamminger, editors, {\em Computer Graphics
  Forum}, volume~2, 2009.

\bibitem{liu_nocedal_lbfgsb}
D.~C. Liu and J.~Nocedal.
\newblock On the limited memory method for large scale optimization.
\newblock {\em Mathematical Programming}, 45:503--528, 1989.

\bibitem{robinette_daanen_paquet_99_caesar}
K.~Robinette, H.~Daanen, and E.~Paquet.
\newblock The {CAESAR} project: A 3-{D} surface anthropometry survey.
\newblock In {\em 3-D Digital Imaging and Modeling}, pages 180--186, 1999.

\bibitem{seo_magnenat-thalmann_03}
H.~Seo and N.~Magnenat-Thalmann.
\newblock An automatic modeling of human bodies from sizing parameters.
\newblock In {\em Proceedings of the 2003 Symposium on Interactive 3D
  Graphics}, pages 19--26, 2003.

\bibitem{seo_etal_shape_from_silhouette}
H.~Seo, Y.~I. Yeo, and K.~Wohn.
\newblock 3d body reconstruction from photos based on range scan.
\newblock {\em Technologies for E-Learning and Digital Entertainment}, pages
  849--860, 2006.

\bibitem{shon_grochow_hertzmann_rao_SGPLVM}
A.~P. Shon, K.~Grochow, A.~Hertzmann, and R.~P.~N. Rao.
\newblock Learning shared latent structure for image synthesis and robotic
  imitation.
\newblock In {\em Neural Information Processing Systems}, 2005.

\bibitem{vanKaick_egstar10}
O.~van Kaick, H.~Zhang, G.~Hamarneh, and D.~Cohen-Or.
\newblock A survey on shape correspondence.
\newblock In {\em Eurographics State-of-the-art Report}, 2010.

\bibitem{Wang2005}
C.~C. Wang.
\newblock Parameterization and parametric design of mannequins.
\newblock {\em Computer Aided Design}, 37:83--98, 2005.

\bibitem{wei_etal_08}
W.~Wei, X.~Luo, and Z.~Li.
\newblock Layer-based mannequin reconstruction and parameterization from 3d
  range data.
\newblock In {\em Advances in Geometric Modeling and Processing}, 2008.

\bibitem{xi_lee_shu_07_bodies}
P.~Xi, W.-S. Lee, and C.~Shu.
\newblock Analysis of segmented human body scans.
\newblock In {\em Graphics Interface}, 2007.

\end{thebibliography}
\end{document}